\documentclass[journal=jcim,manuscript=article]{achemso}

\usepackage[version=3]{mhchem} % Formula subscripts using \ce{}
\usepackage{booktabs}
\usepackage{hyperref}
\author{Junren Li}
\affiliation{BNLMS, College of Chemistry and Molecular Engineering, Peking University, Beijing 100871, China}

\author{Luhua Lai}
\affiliation{BNLMS, College of Chemistry and Molecular Engineering, Peking University, Beijing 100871, China}
\alsoaffiliation{Center for Quantitative Biology, Academy for Advanced Interdisciplinary Studies, Peking University, Beijing 100871, China}
\alsoaffiliation{Peking University Chengdu Academy for Advanced Interdisciplinary Biotechnologies, Chengdu, Sichuan 610213, China}
\alsoaffiliation{Peking-Tsinghua Center for Life Science, Academy for Advanced Interdisciplinary Studies, Peking University, Beijing 100871, China}
\alsoaffiliation{Research Unit of Drug Design Method, Chinese Academy of Medical Sciences (2021RU014), Beijing 100871, China}
\email{lhlai@pku.edu.cn}

\title{SynCraft: Guiding Large Language Models to Predict Edit Sequences for Molecular Synthesizability Optimization}

\abbreviations{IR,NMR,UV}
\keywords{American Chemical Society, \LaTeX}

\begin{document}

% \begin{tocentry}

% Some journals require a graphical entry for the Table of Contents.
% This should be laid out ``print ready'' so that the sizing of the

% Inside the \texttt{tocentry} environment, the font used is Helvetica
% 8\,pt, as required by \emph{Journal of the American Chemical

% The surrounding frame is 9\,cm by 3.5\,cm, which is the maximum
% permitted for  \emph{Journal of the American Chemical Society}
% graphical table of content entries. The box will not resize if the
% content is too big: instead it will overflow the edge of the box.

% This box and the associated title will always be printed on a
% separate page at the end of the document.

% \end{tocentry}

\begin{abstract}
  Generative artificial intelligence has revolutionized the exploration of chemical space, yet a critical bottleneck remains that a substantial fraction of generated molecules is synthetically inaccessible. 
  Current solutions, such as post-hoc filtering or projection-based methods, often compromise structural novelty or disrupt key pharmacophores by forcing molecules into pre-defined synthetic templates. 
  Herein, we introduce SynCraft, a reasoning-based framework that reframes synthesizability optimization not as a sequence translation task, but as a precise structural editing problem. 
  Leveraging the emergent reasoning capabilities of Large Language Models, SynCraft navigates the ``synthesis cliff" where minimal structural modifications yield significant gains in synthetic feasibility. 
  By predicting executable sequences of atom-level edits rather than generating SMILES strings directly, SynCraft circumvents the syntactic fragility of LLMs while harnessing their chemical intuition. 
  Extensive benchmarks demonstrate that SynCraft outperforms state-of-the-art baselines in generating synthesizable analogs with high structural fidelity. 
  Furthermore, through interaction-aware prompting, SynCraft successfully replicates expert medicinal chemistry intuition in editing PLK1 inhibitors and rescuing high-scoring but previously discarded RIPK1 candidates in previous molecular generation literatures. 
  \end{abstract}

\section{Introduction}
The discovery and development of new therapeutics is always a lengthy, costly, and high-risk endeavor\cite{berdigaliyev2020overview}. 
In recent years, artificial intelligence-driven drug discovery (AIDD) has emerged as a transformative paradigm to accelerate this process\cite{sun2025computer, wang2025modeling}. 
One of the key techniques in AIDD is generative models, which have demonstrated a remarkable capacity to explore the vastness of chemical space\cite{tong2021generative, gromski2019explore}. 
The primary objective of these models is often narrowly focused on optimizing a specific property, most notably the predicted binding affinity to a protein target. 
In their pursuit of maximizing this objective function, however, such models frequently operate with an incomplete understanding of chemical principles, generating structures that, while computationally optimal, exhibit poor stability or contain strained, synthetically infeasible motifs. 
This intense focus on a single endpoint, often at the expense of holistic chemical reasonableness, has created a critical bottleneck: a significant portion of the most promising generated hits are practically impossible to synthesize, impeding their translation into tangible laboratory assets\cite{swanson2024generative, papidocha2025elephant, gao2020synthesizability, bilodeau2022generative, van2025search}.

To bridge this gap between virtual design and real-world synthesis, several strategies have been proposed\cite{guo2025directly, sun2025synllama, koziarski2024rgfn}. 
One approach involves post-hoc filtering, where generated candidates are screened using retrosynthesis software, a process that ensures synthesizability but often discards a vast number of potentially valuable molecules\cite{gao2020synthesizability, liu2022retrognn, long2025artificial}
Alternatively, other methods constrain the generation process a priori by constructing molecules exclusively from a finite library of building blocks and reaction templates\cite{yiboli2021structure, wang2022chemistga, yin2025synmolopt, loeffler2024reinvent}, or by projecting a generated molecule onto its nearest analog within such a pre-defined synthesizable space\cite{gao2021synnet, gao2025synformer, lee2025reasyn, chen2025syntwins, luo2025prexsyn}. 
While effective to an extent, these methods impose a fundamental trade-off, which sacrifices the very exploratory power that makes generative models attractive. 
The explorable chemical space is drastically curtailed, and promising scaffolds that lie just outside these rigid boundaries are lost.

However, the distinction between a synthesizable and an unsynthesizable molecule is not always a vast chasm requiring projection into a distant chemical neighborhood. 
We observe that frequently, a minor and localized structural modification is sufficient to transform a challenging target into an accessible one\cite{xie2025transpharmer}. 
This phenomenon is conceptually analogous to the well-established ``activity cliffs" in medicinal chemistry, where a subtle change in a molecular structure can induce a dramatic shift in its biological activity\cite{stumpfe2014recentactivitycliff, stumpfe2012exploringactivitycliff}. 
This leads us to define a corresponding concept, the ``synthesis cliff", where a small, targeted structural edit can dramatically improve synthetic feasibility. 
Therefore, the central goal of this work is to develop a new approach that performs precise, minimal modifications to navigate this synthesis cliff, thereby salvaging promising designs without being confined to a limited repertoire of reaction templates.

The recent advent of Large Language Models (LLMs) offers a compelling new avenue to address this challenge\cite{ramos2025reviewchemllm, zhao2025chemdfmr, zhao2025superchem}. 
Trained on vast corpora of scientific text, these models have developed an extensive, implicit knowledge base of chemical principles, reaction rules, and structural stability. 
Beyond rote knowledge, they exhibit emergent reasoning capabilities to formulate strategies and justify their decisions in human-readable language\cite{wei2022cot}. 
Despite this immense potential, a fundamental obstacle has hindered their direct application to molecular optimization. 
LLMs lack native proficiency in the strict, grammatical formalisms of chemical representations like SMILES\cite{tao2025makevalidsmiles, ganeeva2025measuringllmsmiles, gao2025cidd, zhu2025meco}. 
This syntactic fragility means that when prompted to generate a complete molecule, they often produce invalid or chemically nonsensical outputs, leading to high failure rates and undermining their reliability for scientific tasks.

Herein, we introduce SynCraft, a novel reasoning-based framework designed to harness the chemical intelligence of LLMs while systematically circumventing their syntactic weaknesses. 
Instead of tasking the LLM with the direct generation of a full SMILES string, we reframe its role to that of a chemical strategist. 
Through a Chain-of-Thought (CoT) prompting strategy\cite{wei2022cot}, we guide the model to first reason about the synthetic liabilities within a molecule and then formulate a plan. 
This plan is not a new molecule, but rather a precise sequence of atom-level edits (e.g., DEL\_ATOM, ADD\_BOND). 
These instructions are then passed to a deterministic chemical toolkit for flawless execution, guaranteeing the validity of the final structure\cite{landrum2006rdkit}. 
Through extensive benchmarking, we demonstrate that SynCraft significantly outperforms current state-of-the-art methods, particularly in generating synthesizable analogs with high structural similarity to the original input. 
Moving beyond quantitative similarities, we further illustrate SynCraft’s practical utility in structure-based scenarios, where it functions as an interaction-aware editor to resolve synthetic bottlenecks without disrupting critical binding modes. 
We demonstrate that SynCraft can optimize molecules for synthesizability while maintaining their biological relevance, offering a precise tool for the critical stage of hit optimization.
\begin{figure}[htbp]
\centering
    \includegraphics[width=0.9\textwidth]{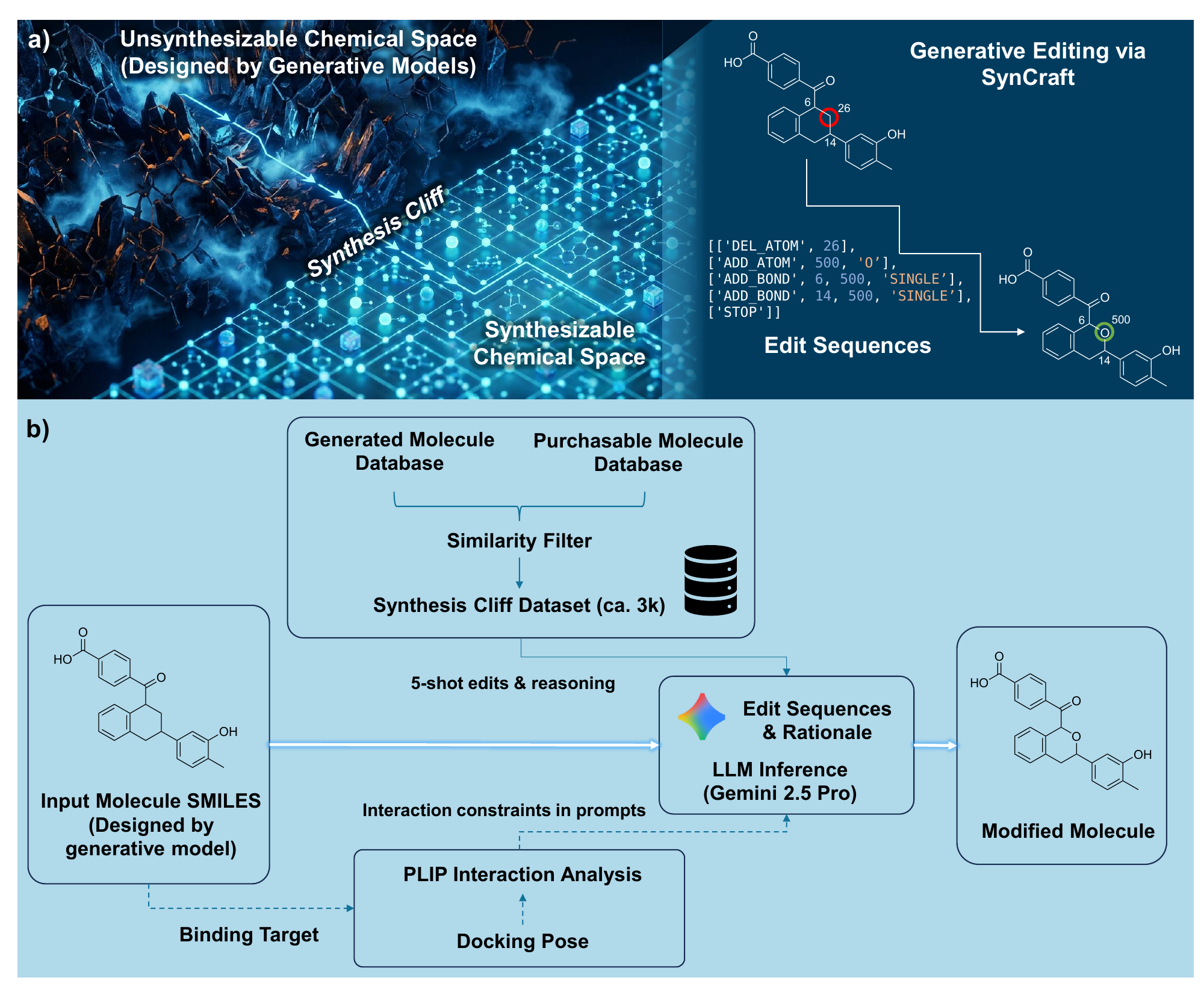}
    \caption{The SynCraft framework.
(a) Concept of the synthesis cliff. Generative models often produce molecules in the ``unsynthesizable highlands." SynCraft navigates to the accessible ``lowlands" not by regenerating the whole molecule, but by predicting a precise sequence of discrete edits (e.g., DEL\_ATOM, ADD\_BOND), transforming the target into a synthesizable analog.
(b) The workflow of SynCraft. The process begins by retrieving similar ``unsynthesizable-to-synthesizable" pairs from a pre-constructed Synthesis Cliff Dataset. These pairs serve as few-shot demonstrations for the Large Language Model (Gemini-2.5-Pro). The model operates via a Chain-of-Thought (CoT) mechanism: first reasoning about synthetic liabilities and (optional) biological constraints derived from PLIP interaction analysis, and then generating an executable edit sequence. This sequence is deterministically applied to the input molecule to yield the final optimized candidate.
    }
    \label{fig:overview}
\end{figure}
\section{Methods}

\subsection{Dataset Construction and Definition of the Synthesis Cliff}

Our investigation commenced with a large and diverse set of virtual molecules designed from structure-based drug discovery campaigns. These molecules were sourced from the GenBench3D benchmark, which evaluates state-of-the-art 3D generative models designed to produce ligands within a protein's binding pocket\cite{baillif2024genbench3d}. We specifically curated the outputs from five prominent models within this benchmark: LIGAN\cite{ragoza2022ligan}, 3D-SBDD\cite{luo20213dsbdd}, Pocket2Mol\cite{peng2022pocket2mol}, DiffSBDD\cite{schneuing2024diffsbdd}, and ResGen\cite{zhang2023resgen}. This initial pool represents a realistic collection of novel chemical entities optimized for predicted binding affinity, often without inherent synthetic constraints, thus providing a challenging testbed for synthesizability correction. The entire collection of generated molecules was first subjected to a rigorous retrosynthetic analysis to assess their synthetic feasibility.

To systematically locate unsynthesizable molecules, we employed the SimpRetro model\cite{li2024simpretro}. Distinguished by its lightweight, CPU-based architecture, SimpRetro enables the cost-effective and high-throughput retrosynthetic analysis of massive datasets (spanning thousands of molecules) without requiring extensive GPU resources. The retrosynthetic search was configured with a comprehensive set of reaction templates derived from the United States Patent and Trademark Office (USPTO) database\cite{lowe2012uspto}, filtered to include all reactions with a reported frequency of two or more. The search for viable starting materials was constrained to the commercially available Enamine building blocks, as the same that are used for training the projection-based models\cite{luo2024projecting, gao2025synformer, lee2025reasyn}. A molecule was operationally defined as ``unsynthesizable" if the SimpRetro algorithm failed to identify a valid synthesis route within a strict time limit of 30 minutes on an Intel Xeon E5-2670 CPU, with no maximum step limit imposed on the search.

From the pool of molecules designated as unsynthesizable, we constructed our ``Synthesis Cliff" dataset, which provides the ground-truth examples for our study. To do this, we first randomly sampled 4,000 unsynthesizable molecules from each of the five generative models. For each of these query molecules, we then performed a large-scale similarity search against the eMolecules database, a comprehensive collection of approximately 23 million commercially available compounds. A successful ``cliff pair" was formed by identifying the nearest neighbor in the eMolecules database that satisfied two criteria: a ECFP4 Tanimoto similarity greater than 0.5\cite{rogers2010ecfp} and a 2D pharmacophore similarity (Pharm2D)\cite{mcgregor1999pharm2d} greater than 0.5. This procedure resulted in our final dataset of 3,332 pairs, each consisting of an unsynthesizable molecule and its closest, structurally similar, commercially available counterpart. We selected the similarity threshold of 0.5 to balance data sufficiency with structural diversity. This threshold ensures a robust volume of reference pairs for few-shot retrieval while remaining permissive enough to capture non-trivial structural transformations (e.g., scaffold hopping).

The remaining unsynthesizable molecules, which were not sampled for the cliff pair construction, constituted the final test set used for evaluating all methods. An initial analysis revealed that molecules generated by models such as LIGAN and 3D-SBDD had exceptionally low rates of synthetic accessibility. Consequently, to ensure a meaningful and challenging evaluation, the results and discussion presented in this work primarily focus on the test sets derived from the more reliable ResGen and Pocket2Mol models, resulting in 1,025 and 1,725 molecules, respectively.

\subsection{The SynCraft Framework: Generative Editing via In-Context Reasoning}

SynCraft reframes synthesizability optimization not as a sequence translation task, but as a discrete structure editing problem. By decoupling strategic planning from chemical execution, we utilize a LLM (Gemini-2.5-Pro\cite{team2023gemini}) to predict precise modification instructions, which can be deterministically applied to the molecular graph.

\subsubsection{Discrete Editing Action Space}
To enable precise control over molecular topology, we defined a compact action space $\mathcal{A}$ where operations are referenced via unique atom-map numbers. The LLM is restricted to outputting a JSON-formatted sequence of commands. The supported operations are:

\begin{itemize}
    \item \texttt{DEL\_ATOM}: Removes a specific atom and its incidental bonds based on the map number.
    \item \texttt{MUTATE\_ATOM}: Changes the atomic element of a targeted atom (e.g., C $\to$ N).
    \item \texttt{ADD\_ATOM}: Introduces a new atom with a unique ID (indexed $\geq 500$) to distinguish it from the original scaffold.
    \item \texttt{ADD\_BOND} / \texttt{DEL\_BOND}: Creates or removes a bond between atoms, with explicit definition of bond order.
    \item \texttt{CHANGE\_BOND}: Modifies the order or aromaticity of an existing bond.
    \item \texttt{SET\_CHIRAL} / \texttt{SET\_BOND\_STEREO}: Explicitly defines stereocenters or E/Z isomerism to resolve ambiguity.
\end{itemize}

\subsubsection{Constructing the Reference Dataset}
To ground the model's generation in chemical reality, we transformed the ``Synthesis Cliff" dataset into a corpus of instructional examples. This process involved two steps: edit extraction and rationale synthesis.

For each pair of unsynthesizable source ($M_{src}$) and synthesizable target ($M_{tgt}$) molecules, we employed a graph matching algorithm to extract the minimal edit distance path. This converted the structural differences into a ground-truth sequence of actions from space $\mathcal{A}$.

Since raw edit sequences lack semantic intent, we augmented this dataset with ``ground truth". We fed the $(M_{src}, M_{tgt})$ pairs into Gemini-2.5-Pro, instructing it to analyze the transformation post-hoc and generate a medicinal chemistry rationale (e.g., ``removing a chiral center to avoid stereoisomeric mixtures"). This resulted in a knowledge base of references, each containing a pair of molecules, a coherent chemical rationale, and the correct executable edit sequence.

\subsubsection{Retrieval-Augmented Inference}
During the inference phase, SynCraft operates via a Retrieval-Augmented Generation (RAG) paradigm\cite{brown2020fewshotlearning, lewis2020retrieval}. For a given query molecule, we perform a similarity search against the reference database using Morgan fingerprints (radius 2, 2048-bit) to retrieve the top-$k$ (typically $k=5$) most structurally similar golden examples.

These retrieved examples serve as few-shot demonstrations in the system prompt. By exposing the model to these ``Reasoning $\to$ Edit" exemplars, we enforce a Chain-of-Thought (CoT) workflow\cite{wei2022cot}. The model is strictly instructed to first articulate a natural language analysis identifying specific synthetic liabilities, and only then generate the JSON edit sequence. This mechanism significantly reduces hallucination and ensures that modifications are chemically justified. The full system prompts, along with examples of the retrieved few-shot exemplars including their reasoning traces and JSON outputs, are illustrated in Supporting Information Section S1.3.

\subsection{Interaction-Aware Prompting Strategy}

A critical challenge in synthesizability optimization is avoiding the inadvertent removal of key pharmacophores. SynCraft addresses this by embedding a structure-aware constraint mechanism that translates 3D protein-ligand interaction data into semantic constraints within the LLM's context window.

Specifically, for each input molecule, we execute an automated profiling pipeline. The molecule is first converted to a 3D conformer and docked into the target protein pocket using \texttt{AutoDock Vina}\cite{trott2010vina}. The optimal pose is then analyzed using \texttt{PLIP} to categorize non-covalent interactions, such as hydrogen bonds and $\pi$-stacking\cite{salentin2015plip, schake2025plip2025}. A pivotal step in this pipeline is \textbf{graph alignment}: since docking software often re-indexes atoms, we implement a substructure matching algorithm to map the atom indices from the 3D docked pose back to the canonical atom-map numbers used in the 2D editing action space. This ensures that spatial constraints are precisely localized to the correct nodes in the molecular graph.

To make these 3D constraints comprehensible to the LLM, we transform the aligned interaction profile into natural language descriptions. For each critical atom, a descriptive prompt is generated following the template:
\begin{equation}
\label{eq:prompt}
\text{``Atom } [i] \text{ (} \text{Element } E \text{, connected to } [N]) \text{: } \mathcal{D}_{interaction} \text{."}
\end{equation}
where $[i]$ is the atom map number, $[N]$ denotes neighboring atoms for local context, and $\mathcal{D}_{interaction}$ describes the biological role (e.g., \textit{``forms a critical Hydrogen Bond with LYS43"}). These descriptions are injected into the system prompt under a \texttt{[CRITICAL BIOLOGICAL CONSTRAINTS]} header. The LLM is explicitly instructed to preserve these atoms or propose bioisosteric replacements, thereby transforming the task into a constrained optimization problem that balances synthetic feasibility with biological activity. Please refer to Supporting Information Section S1.4 for the full prompt templates.
\section{Results and discussion}
\subsection{Quantitative Evaluation of Synthesizability Optimization}

We first evaluated the capacity of SynCraft to navigate the ``synthesis cliff" on a large scale. As described in the Methods section, our test set comprises molecules generated by two state-of-the-art structure-based models, Pocket2Mol (P2M) and ResGen, which were initially identified as unsynthesizable by SimpRetro. We benchmarked SynCraft against three baselines: ChemProjector\cite{luo2024projecting}, SynFormer\cite{gao2025synformer}, and ReaSyn\cite{lee2025reasyn}. The primary metric is the success rate of finding a synthesizable analog that maintains a Tanimoto similarity (calculated via ECFP4 with a radius of 2, 4096 bits) above a specified threshold relative to the original generated molecule.

It is important to note that, during this test, SynCraft was deployed in its standard mode without the interaction-aware constraint module. This experimental design was chosen to isolate and evaluate the model's intrinsic ability to restore synthesizability based solely on chemical reasoning and structural fidelity. Furthermore, our baselines did not explicitly incorporate the protein structure when projecting unsynthesizable molecules. Consequently, we focus here on geometric and synthetic metrics, while the critical assessment of biological activity preservation is reserved for the detailed case studies presented later.

To ensure a comprehensive comparison, we allowed the projection-based baselines to generate unlimited candidates according to their default settings, selecting the highest-similarity synthesizable molecule as the final output. In contrast, SynCraft was restricted to predicting only five parallel edit sequences. The quantitative results are summarized in Table \ref{tab:quant_results}. SynCraft demonstrates a substantial improvement over all baselines across the majority of similarity thresholds, particularly in the challenging regime where significant structural edits are required to restore synthesizability.

\begin{table}[ht]
\centering
\caption{Success rates of identifying synthesizable analogs at varying Tanimoto similarity thresholds. The test sets consist of unsynthesizable molecules generated by Pocket2Mol (P2M) and ResGen. Best results are highlighted in \textbf{bold}.}
\label{tab:quant_results}
\resizebox{\textwidth}{!}{%
\begin{tabular}{lcccc}
\toprule
\textbf{Method} & \textbf{Sim $>$ 0.5} & \textbf{Sim $>$ 0.6} & \textbf{Sim $>$ 0.7} & \textbf{Sim $>$ 0.8} \\
\midrule
\multicolumn{5}{c}{\textit{Source Model: Pocket2Mol (P2M)}} \\
\midrule
P2M-ChemProjector & 21.6\% & 9.8\% & 4.2\% & 1.9\% \\
P2M-SynFormer & 30.7\% & 14.1\% & 6.2\% & 2.9\% \\
P2M-ReaSyn & 30.4\% & 15.4\% & 6.4\% & 3.0\% \\
\textbf{P2M-SynCraft (Ours)} & \textbf{42.7\%} & \textbf{28.4\%} & \textbf{11.9\%} & \textbf{3.2\%} \\
\midrule
\multicolumn{5}{c}{\textit{Source Model: ResGen}} \\
\midrule
ResGen-ChemProjector & 28.3\% & 14.8\% & 7.2\% & 2.0\% \\
ResGen-SynFormer & 33.8\% & 19.9\% & 8.7\% & 3.5\% \\
ResGen-ReaSyn & 37.5\% & 22.0\% & 10.5\% & \textbf{4.3\%} \\
\textbf{ResGen-SynCraft (Ours)} & \textbf{44.7\%} & \textbf{29.1\%} & \textbf{13.1\%} & 3.5\% \\
\bottomrule
\end{tabular}%
}
\end{table}

On the Pocket2Mol dataset, SynCraft achieves a success rate of 42.7\% at a similarity threshold of 0.5, outperforming the strongest baseline (SynFormer/ReaSyn, $\approx$30\%) by a margin of over 12 percentage points. This advantage is even more pronounced at the strict threshold of 0.6, where SynCraft (28.4\%) nearly doubles the success rate of SynFormer (14.1\%) and ReaSyn (15.4\%). A similar trend is observed with the ResGen dataset, where SynCraft maintains a dominant lead at lower and medium similarity thresholds (44.7\% at $>$0.5 and 29.1\% at $>$0.6).

These results underscore the limitations of projection-based methods when the ``synthesis cliff" requires precise structural modifications. Projection methods often struggle to find a valid pathway if the nearest synthesizable neighbor lies outside their achievable chemical space. In contrast, SynCraft, leveraging the reasoning capabilities of LLMs, can perform flexible, atom-level editing (e.g., ring opening, heteroatom substitution) to navigate the landscape more effectively.

To validate our architectural choices, we performed extensive ablation studies comparing ``edit prediction" versus ``direct SMILES generation", as well as varying the number of few-shot examples. The results demonstrate that the edit-based paradigm significantly reduces structural hallucination and outperforms direct generation methods, particularly at high similarity thresholds. Detailed data and discussions on these ablation experiments are presented in Supporting Information Section S2.

\subsection{SynCraft Navigates the Synthesis Cliff via Bioisosteric Editing while Preserving Pharmacophores}
\begin{figure}[htbp]
\centering
    \includegraphics[width=\textwidth]{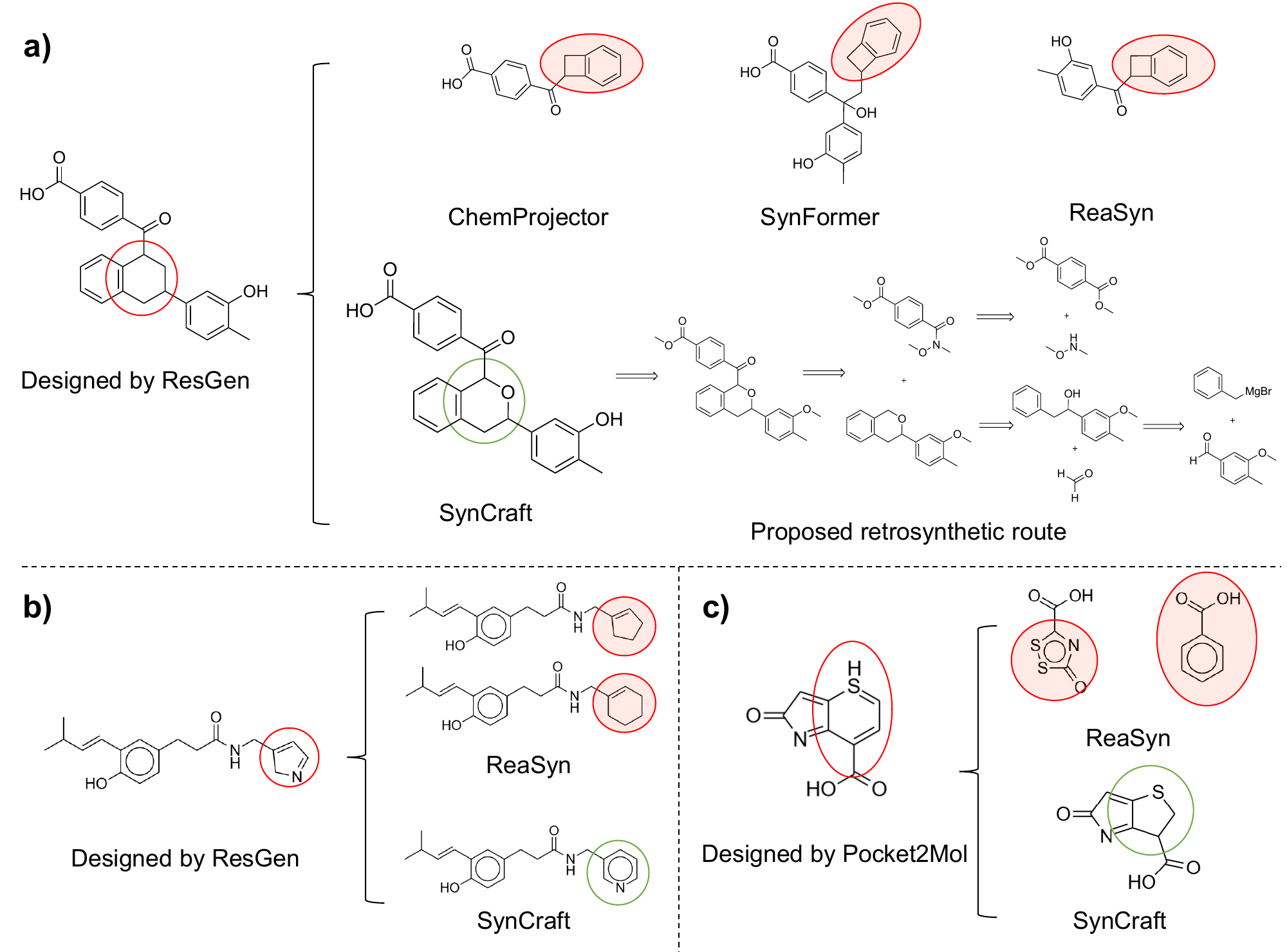}
    \caption{Qualitative comparison of structural modifications by SynCraft versus projection-based baselines. (a) Navigating the synthesis cliff via bioisosteric editing. The input molecule features a complex fused carbocycle. SynCraft (bottom path) strategically replaces a methylene group with an oxygen atom, creating a synthetically accessible ether linkage while preserving the scaffold shape. In contrast, projection-based baselines converge on a strained benzocyclobutene-like motif, likely an artifact of projecting into a sparse region of the synthesizable chemical space.
    (b, c) Preserving pharmacophoric integrity against oversimplification. (b) For a nitrogen-containing ring, the baseline (ReaSyn) discards the nitrogen atom, losing a potential interaction site. SynCraft aromatizes the ring to a pyridine derivative, stabilizing the structure while preserving the heteroatom. (c) For a fused bicyclic core, the baseline breaks the scaffold into simpler fragments (scaffold hopping), potentially disrupting shape complementarity. 
    }
    \label{fig:syn_vs_proj}
\end{figure}
To better understand the distinct behaviors of SynCraft compared to projection-based baselines, we visually inspected the structural modifications proposed by different models.

\subsubsection{Navigate the Synthesis Cliff via Bioisosteric Editing}

A key advantage of SynCraft is its ability to perform ``surgical" edits that resolve synthetic bottlenecks while minimally perturbing the molecular scaffold. A representative example is shown in Figure \ref{fig:syn_vs_proj}. The input molecule features a fused bicyclic carbocycle, a motif that can be challenging to construct with specific substitution patterns.

SynCraft navigates this ``synthesis cliff" by identifying a precise, single-atom modification: replacing a methylene carbon ($-\text{CH}_2-$) in the saturated ring with an oxygen atom ($-\text{O}-$) (Figure \ref{fig:syn_vs_proj}A). This edit converts the carbocycle into a chroman-like derivative. From a synthetic perspective, this is a highly strategic transformation. It unlocks a robust retrosynthetic disconnection via C-O bond formation, which is generally more accessible than constructing the corresponding C-C bonds in the all-carbon analog. Please refer to the proposed retrosynthetic route in Figure~\ref{fig:syn_vs_proj}A for more details.
% Importantly, this C$\to$O exchange acts as a bioisosteric replacement, likely preserving the overall shape and physicochemical profile of the original design.

In contrast, projection-based baselines operate by mapping the input to the nearest synthesizable neighbor in a predefined space, yielding a different outcome for the same input. As shown in Figure \ref{fig:syn_vs_proj}A, these models converged on a structure containing a benzocyclobutene-like fused four-membered ring. While this motif exists in chemical databases, its formation here likely represents an artifact of the projection process—where the model, unable to find a close match for the specific 6,6-fused system in its building block library, snaps to a topologically distinct neighbor. This highlights a fundamental distinction: SynCraft modifies the molecule to \textit{enable} synthesis, whereas projection models search for a synthesizable \textit{proxy}, which may occasionally lead to strained or chemically idiosyncratic motifs when the local chemical space is sparse.

\subsubsection{Preserving Pharmacophoric Integrity}

Another critical consideration is the preservation of key pharmacophores during optimization. A potential risk in automated synthesizability correction is the oversimplification of molecular scaffolds, where complex but essential features are removed to satisfy synthetic scores.

Figure \ref{fig:syn_vs_proj} illustrates two scenarios where SynCraft demonstrates superior pharmacophore retention:
\begin{itemize}
    \item \textbf{Heteroatom Preservation:} In the first case (Figure \ref{fig:syn_vs_proj}B), the input contains a functionalized nitrogen-containing ring. The baseline method (ReaSyn) proposes a solution that discards the nitrogen atom and saturates the ring, resulting in a simple cycloalkane. While synthetically trivial, this removes a potential hydrogen bond acceptor/donor site. SynCraft, conversely, resolves the synthetic liability by aromatizing the ring into a pyridine derivative. This edit stabilizes the structure while strictly preserving the nitrogen atom, maintaining the potential for polar interactions.
    \item \textbf{Scaffold Consistency:} In the second case (Figure \ref{fig:syn_vs_proj}C), the input features a fused bicyclic core. The baseline resolves the synthetic difficulty by breaking the fused system into simpler, acyclic fragments. Such ``scaffold hopping" risks destroying the 3D shape complementarity required for binding. SynCraft leverages its editing capabilities to preserve the fused topology, applying minimal bond order adjustments to ensure valency rules are met without dismantling the core skeleton.
\end{itemize}

\begin{figure}[htbp]
\centering
    \includegraphics[width=\textwidth]{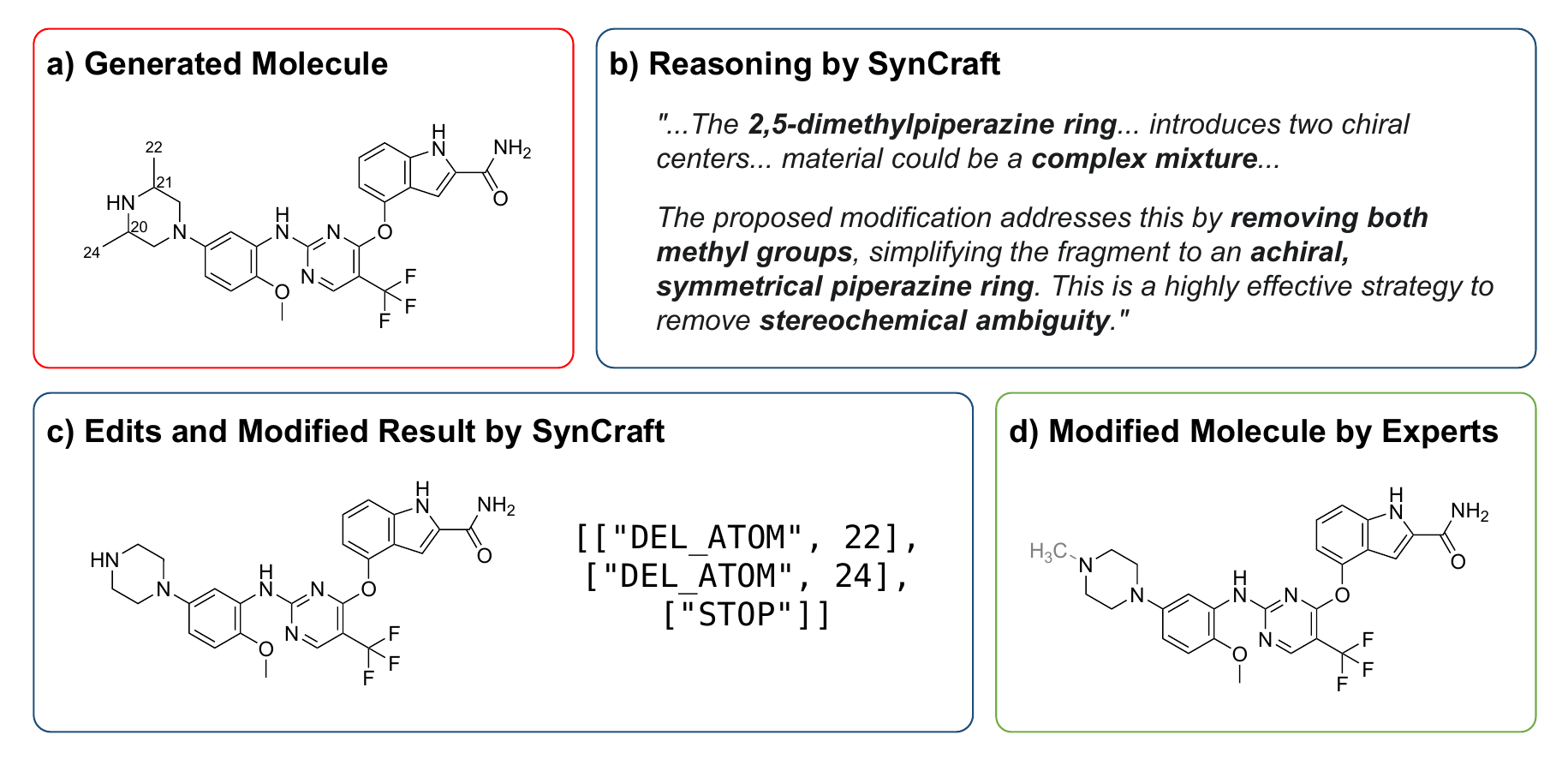}
    \caption{Retrospective validation of SynCraft’s reasoning capability against human expert intuition in TransPharmer\cite{xie2025transpharmer}.
(a) The initial generated molecule (lig-886) contains a 2,5-dimethylpiperazine ring, which introduces two chiral centers and poses a risk of forming complex stereoisomeric mixtures.
(b) SynCraft correctly identifies this synthetic liability in its reasoning trace, explicitly stating the need to remove stereochemical ambiguity.
(c) The edit sequence generated by SynCraft removes the two methyl groups on the carbon skeleton.
(d) The actual chemically synthesized lead compound from the original study (IIP0944). SynCraft’s decision to simplify the core to an achiral piperazine converges precisely with the strategy adopted by human experts to ensure developability.
    }
    \label{fig:retro_case}
\end{figure}
These qualitative comparisons suggest that SynCraft acts akin to a ``molecular editor", applying local fixes to ensure validity. 
% , whereas projection methods often function as "nearest-neighbor retrievers," which can be highly effective but are constrained by the density of the underlying template library.

\subsection{Retrospective Validation: Replicating Expert Intuition in Hit Optimization}

To further validate whether SynCraft's editing logic aligns with human medicinal chemistry expertise, we conducted a retrospective case study using a recently published work on pharmacophore-based generation\cite{xie2025transpharmer}. In that study, the authors generated a series of PLK1 inhibitors but noted that the initially generated structures often required manual modification by chemists to ensure synthesizability before wet-lab testing. This ``AI-generation $\to$ Human-modification" workflow provides a perfect ground truth to test if SynCraft can autonomously replicate the ``Human-modification" step.

We focused on the transformation from the generated candidate \texttt{lig-886} to the synthesized compound \texttt{IIP0944} (Figure \ref{fig:retro_case}). The initially generated molecule, \texttt{lig-886}, contains a 2,5-dimethylpiperazine moiety. From a synthetic perspective, this substructure is a significant liability: it introduces two chiral centers, potentially yielding a complex mixture of up to four stereoisomers (cis/trans enantiomeric pairs). Synthesizing a single pure stereoisomer requires expensive chiral resolution or asymmetric synthesis, which is often unjustifiable for an early-stage hit.

When \texttt{lig-886} was fed into SynCraft with interaction-awareness (PDB ID: 2YAC), our model successfully identified this bottleneck. The generated rationale explicitly stated: \textit{``The primary synthetic liability... is the 2,5-dimethylpiperazine ring... producing the compound as a mixture of isomers is undesirable... The proposed modification addresses this by removing both methyl groups... eliminating complex stereoisomeric issues."}

Guided by this reasoning, SynCraft executed the editing plan to remove the steric carbon-methyl groups, resulting in a simplified, achiral piperazine scaffold. Remarkably, this autonomous decision converges precisely with the strategy adopted by the human experts in the original study. The chemically synthesized lead, \texttt{IIP0944}, shares the exact same de-methylated piperazine core. Although the human chemists additionally employed an $N$-methyl group (likely to address potential metabolic instability or using $N$-methylpiperazine as a readily available reagent), SynCraft's core decision—to resolve the ``synthesis cliff" by removing the problematic chiral carbons while preserving the key nitrogen pharmacophore, mirroring expert intuition.

This case demonstrates that SynCraft goes beyond simple rule-matching; it exhibits a ``molecular reasoning" capability that can identify and resolve subtle stereochemical liabilities that typically require the judgment of experienced medicinal chemists.

\subsection{Prospective Rescue of Shelved Candidates: Interaction-Aware Optimization of High-Scoring Designs}

\begin{figure}[htbp]
\centering
    \includegraphics[width=0.7\textwidth]{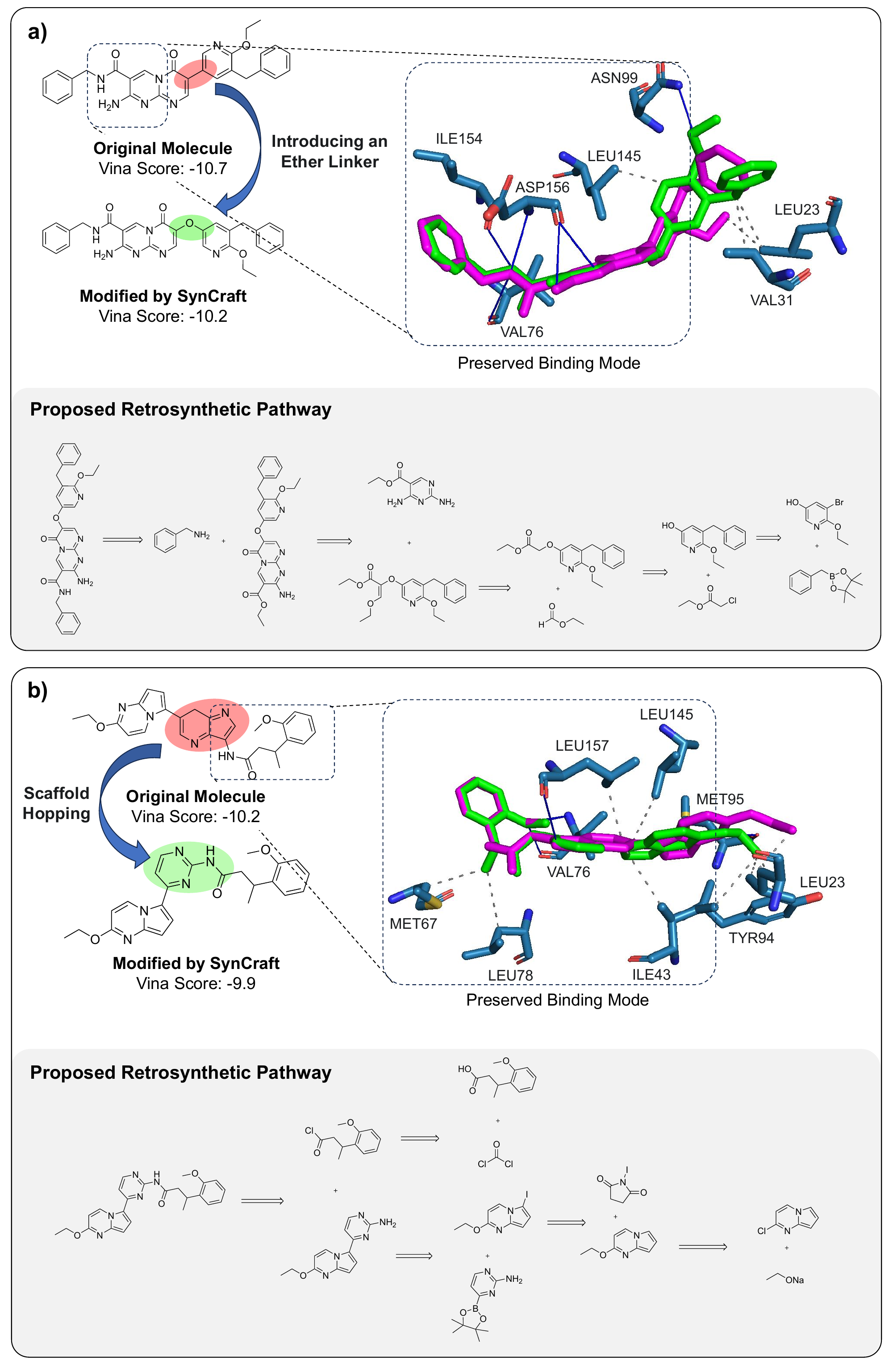}
    \caption{Prospective rescue of shelved high-scoring drug candidates targeting RIPK1.
(a) Case I: SynCraft introduces an ether linker to replace a challenging C-C bond between two electron-deficient aromatic rings. This modification facilitates synthesis via C-O bond formation (proposed retrosynthetic pathway shown below) while maintaining the binding mode (Vina score: -10.7 to -10.2 kcal/mol.
(b) Case II: SynCraft replaces a complex, non-planar fused polycyclic system containing a quaternary carbon with a planar, synthetically modular 2,4-disubstituted pyrimidine scaffold (Vina score: -10.2 to -9.9 kcal/mol).
\textit{Green sticks: Original molecule; Magenta sticks: SynCraft-optimized molecule; Blue lines: Hydrogen bonds; Grey dashed lines: Hydrophobic interactions.
    }}
    \label{fig:rescue_cases}
\end{figure}

In generative drug design campaigns, a common tragedy is the ``shelving" of high-scoring candidates. For instance, in the recent discovery of RIPK1 inhibitors, the authors generated thousands of molecules and prioritized them using a docking-based reward function. However, due to practical resource constraints and synthetic complexity, only a very small portion of the top-ranking molecules were synthesized (8 out of the top 50). This leaves a vast number of potentially potent but synthetically challenging ``orphan" designs\cite{li2022ripk}.

We applied SynCraft to the set of 42 high-scoring but synthetically challenging candidates. To demonstrate the model's versatility, we detail two representative cases below that highlight distinct optimization strategies in Figure~\ref{fig:rescue_cases}. The complete optimization results for all the 42 candidates, including their generated edit sequences and optimized structures, are provided in Supporting Information Section S3. During this process, we tasked SynCraft to optimize them using the interaction-aware prompting strategy (PDB ID: 7YDX), aiming to restore synthesizability while maintaining their predicted binding affinity (Vina score) and key interaction fingerprints.

\subsubsection{Case I: Ether Bridge Replacement of Challenging Biaryl Linkages}

The first candidate (Figure \ref{fig:rescue_cases}A, upper left) exhibited a strong docking score (-10.7 kcal/mol). However, it contained a direct C-C bond between the pyrazolo[3,4-d]pyrimidinone core and a pyridine ring. SynCraft's reasoning engine correctly identified this as the primary bottleneck: \textit{``This bond connects two electron-deficient, sterically hindered aromatic systems, making its formation via standard cross-coupling reactions (like Suzuki or Stille) exceptionally difficult and likely low-yielding."}

Instead of discarding the scaffold, SynCraft proposed inserting an oxygen atom to form an ether linkage. The model reasoned that this modification \textit{``converts the challenging C-C cross-coupling into a much more reliable C-O bond formation."}.

Importantly, this modification was not a random edit but a structure-aware decision. Analysis of the binding pose (Figure \ref{fig:rescue_cases}A, right) confirms that the ether oxygen acts as a spacer that mimics the geometry of the original biaryl bond. The optimized molecule (-10.2 kcal/mol) preserves the critical hydrogen bond network with the hinge region (Asp156) and maintains the hydrophobic contacts with Leu23. The slight flexibility introduced by the ether linker allows the pyridine ring to adjust its orientation to minimize steric clash while retaining the key hydrophobic interactions. 

\subsubsection{Case II: Scaffold Hopping from Intractable Fused Systems to Accessible Heterocycles}

The second candidate (Figure \ref{fig:rescue_cases}B, upper left) featured an exotic, non-planar fused polycyclic core (diazepino[1,2-f]purine-like) containing an isolated saturated methylene unit. While this rigid core fit the pocket well (-10.2 kcal/mol), it represented a ``synthetic nightmare." SynCraft critiqued this structure as having \textit{``a significant synthetic liability... the sheer number of conformationally locked centers hinders large-scale production."}

In a bold move of scaffold hopping, SynCraft replaced the entire fused core with a planar, 2,4-disubstituted pyrimidine scaffold. The model justified this simplification by noting that the pyrimidine ring is \textit{``readily accessible through sequential nucleophilic aromatic substitution,"} offering a modular assembly path.

Despite the drastic reduction in structural complexity, the interactions were successfully preserved. The model explicitly aligned the N1 atom of the new pyrimidine ring to mimic the H-bond accepting role of the original N12 atom. The substituents were re-attached at the C2 and C4 positions to maintain the spatial vectors of the imidazopyridine moiety and the amide side chain. The resulting molecule achieved a comparable docking score of -9.9 kcal/mol. This case exemplifies SynCraft's ability to act as a ``molecular architect," capable of identifying the pharmacophoric essence of a complex design and transplanting it onto a synthetically tractable skeleton.

Collectively, these cases illustrate that SynCraft can effectively ``resurrect" shelved designs. By providing explicit chemical reasoning and executable edit sequences, it transforms computationally promising but practically dead ends into actionable hit compounds. 
We would also like to emphasize that SynCraft is envisioned primarily as a high-precision tool for high-value hit candidates. Based on current API pricing, the operational cost is estimated to be approximately \$3--5 USD per 100 optimization runs. While highly effective, the inference costs associated with LLM reasoning currently make SynCraft less suitable for the mass enumeration of massive virtual libraries from scratch. 
For such large-scale, low-fidelity expansion tasks, traditional template-based methods remain a cost-effective alternative, whereas SynCraft shines in the critical stage of hit optimization where chemical precision outweighs computational cost.

\section{Conclusion}

The divide between virtual chemical space and physical reality has long hindered the translation of AI-designed molecules into laboratory assets. 
In this work, we present SynCraft, a methodology that bridges this gap by treating synthesizability optimization as a strategic reasoning task rather than a pattern-matching problem.
By decoupling strategic planning (via LLM reasoning) from chemical execution (via deterministic editing), SynCraft addresses the twin challenges of syntactic validity and semantic preservation that plague traditional generative models. 
Our results highlight three key advances: (1) the ability to navigate the ``synthesis cliff" with surgical precision, achieving higher success rates than projection-based methods; (2) the capacity to replicate the intuition of medicinal chemists, as evidenced by the retrospective validation on PLK1 inhibitors; and (3) the potential to rescue valuable but synthetically complex ``orphan" designs through interaction-aware scaffold hopping.
Unlike ``black box" models, SynCraft provides human-readable rationales for its modifications, fostering trust and collaboration between AI agents and human chemists. 
As LLMs continue to evolve, we anticipate that such reasoning-centric frameworks will become indispensable tools in the drug discovery pipeline, enabling the automated design of molecules that are not only potent and novel but also readily manufacturable.

\section{Acknowledgment}

This work is supported by the National Natural Science Foundation of China (22237002, T2321001) and the CAMS Innovation Fund for Medical Sciences (2021-I2M-5-014). The authors thank Prof. Zhirong Liu, Yuzhe Wang, Zihan Wang and Zhixian Huang for helpful discussion.

\section{Data \& Code Availability}
The source code for SynCraft, including the synthesis cliff data, prompt templates, and the editing pipeline, is available on GitHub at \url{https://github.com/catalystforyou/SynCraft-Core}.

\begin{suppinfo}

Details on the interaction-aware constraint pipeline, extensive ablation studies and the other revised molecule results are available in the Supporting Information.

\end{suppinfo}

\bibliography{achemso-demo}

@article{berdigaliyev2020overview,
  title={An overview of drug discovery and development},
  author={Berdigaliyev, Nurken and Aljofan, Mohamad},
  journal={Future medicinal chemistry},
  volume={12},
  number={10},
  pages={939--947},
  year={2020},
  publisher={Taylor \& Francis}
}

@article{sun2025computer,
  title={Computer-Aided Drug Discovery for Undruggable Targets},
  author={Sun, Qi and Wang, Hanping and Xie, Juan and Wang, Liying and Mu, Junxi and Li, Junren and Ren, Yuhao and Lai, Luhua},
  journal={Chemical Reviews},
  year={2025},
  publisher={ACS Publications}
}

@article{wang2025modeling,
  title={Modeling protein--ligand interactions for drug discovery in the era of deep learning},
  author={Wang, Yuzhe and Li, Yibo and Chen, Jiaxiao and Lai, Luhua},
  journal={Chemical Society Reviews},
  year={2025},
  publisher={Royal Society of Chemistry}
}

@article{tong2021generative,
  title={Generative models for de novo drug design},
  author={Tong, Xiaochu and Liu, Xiaohong and Tan, Xiaoqin and Li, Xutong and Jiang, Jiaxin and Xiong, Zhaoping and Xu, Tingyang and Jiang, Hualiang and Qiao, Nan and Zheng, Mingyue},
  journal={Journal of Medicinal Chemistry},
  volume={64},
  number={19},
  pages={14011--14027},
  year={2021},
  publisher={ACS Publications}
}

@article{gromski2019explore,
  title={How to explore chemical space using algorithms and automation},
  author={Gromski, Piotr S and Henson, Alon B and Granda, Jaros{\l}aw M and Cronin, Leroy},
  journal={Nature Reviews Chemistry},
  volume={3},
  number={2},
  pages={119--128},
  year={2019},
  publisher={Nature Publishing Group UK London}
}

@article{swanson2024generative,
  title={Generative AI for designing and validating easily synthesizable and structurally novel antibiotics},
  author={Swanson, Kyle and Liu, Gary and Catacutan, Denise B and Arnold, Autumn and Zou, James and Stokes, Jonathan M},
  journal={Nature machine intelligence},
  volume={6},
  number={3},
  pages={338--353},
  year={2024},
  publisher={Nature Publishing Group UK London}
}

@article{papidocha2025elephant,
  title={The elephant in the lab: Synthesizability in generative small-molecule design},
  author={Papidocha, Sven Michael and Burger, Andreas and Bernales, Varinia and Aspuru-Guzik, Al{\'a}n},
  year={2025}
}

@article{gao2020synthesizability,
  title={The synthesizability of molecules proposed by generative models},
  author={Gao, Wenhao and Coley, Connor W},
  journal={Journal of chemical information and modeling},
  volume={60},
  number={12},
  pages={5714--5723},
  year={2020},
  publisher={ACS Publications}
}

@article{bilodeau2022generative,
  title={Generative models for molecular discovery: Recent advances and challenges},
  author={Bilodeau, Camille and Jin, Wengong and Jaakkola, Tommi and Barzilay, Regina and Jensen, Klavs F},
  journal={Wiley Interdisciplinary Reviews: Computational Molecular Science},
  volume={12},
  number={5},
  pages={e1608},
  year={2022},
  publisher={Wiley Online Library}
}

@article{van2025search,
  title={In search of beautiful molecules: a perspective on generative modeling for drug design},
  author={van den Broek, Remco L and Patel, Shivam and van Westen, Gerard JP and Jespers, Willem and Sherman, Woody},
  journal={Journal of chemical information and modeling},
  volume={65},
  number={18},
  pages={9383--9397},
  year={2025},
  publisher={ACS Publications}
}

@article{guo2025directly,
  title={Directly optimizing for synthesizability in generative molecular design using retrosynthesis models},
  author={Guo, Jeff and Schwaller, Philippe},
  journal={Chemical Science},
  volume={16},
  number={16},
  pages={6943--6956},
  year={2025},
  publisher={Royal Society of Chemistry}
}

@article{chen2025syntwins,
  title={SynTwins: a retrosynthesis-guided framework for synthesizable molecular analog generation},
  author={Chen, Shuan and Nam, Gunwook and Aspuru-Guzik, Al{\'a}n and Jung, Yousung},
  journal={Chemical Science},
  year={2025},
  publisher={Royal Society of Chemistry}
}

@article{loeffler2024reinvent,
  title={Reinvent 4: modern AI--driven generative molecule design},
  author={Loeffler, Hannes H and He, Jiazhen and Tibo, Alessandro and Janet, Jon Paul and Voronov, Alexey and Mervin, Lewis H and Engkvist, Ola},
  journal={Journal of Cheminformatics},
  volume={16},
  number={1},
  pages={20},
  year={2024},
  publisher={Springer}
}

@article{yin2025synmolopt,
  title={Syn-MolOpt: a synthesis planning-driven molecular optimization method using data-derived functional reaction templates},
  author={Yin, Xiaodan and Wang, Xiaorui and Wu, Zhenxing and Li, Qin and Kang, Yu and Deng, Yafeng and Luo, Pei and Liu, Huanxiang and Shi, Guqin and Wang, Zheng and others},
  journal={Journal of Cheminformatics},
  volume={17},
  number={1},
  pages={27},
  year={2025},
  publisher={Springer}
}

@article{wang2022chemistga,
  title={ChemistGA: a chemical synthesizable accessible molecular generation algorithm for real-world drug discovery},
  author={Wang, Jike and Wang, Xiaorui and Sun, Huiyong and Wang, Mingyang and Zeng, Yundian and Jiang, Dejun and Wu, Zhenxing and Liu, Zeyi and Liao, Ben and Yao, Xiaojun and others},
  journal={Journal of medicinal chemistry},
  volume={65},
  number={18},
  pages={12482--12496},
  year={2022},
  publisher={ACS Publications}
}

@article{yiboli2021structure,
  title={Structure-based de novo drug design using 3D deep generative models},
  author={Li, Yibo and Pei, Jianfeng and Lai, Luhua},
  journal={Chemical science},
  volume={12},
  number={41},
  pages={13664--13675},
  year={2021},
  publisher={Royal Society of Chemistry}
}

@article{koziarski2024rgfn,
  title={Rgfn: Synthesizable molecular generation using gflownets},
  author={Koziarski, Micha{\l} and Rekesh, Andrei and Shevchuk, Dmytro and van der Sloot, Almer and Gai{\'n}ski, Piotr and Bengio, Yoshua and Liu, Chenghao and Tyers, Mike and Batey, Robert},
  journal={Advances in Neural Information Processing Systems},
  volume={37},
  pages={46908--46955},
  year={2024}
}

@article{liu2022retrognn,
  title={RetroGNN: fast estimation of synthesizability for virtual screening and de novo design by learning from slow retrosynthesis software},
  author={Liu, Cheng-Hao and Korablyov, Maksym and Jastrzebski, Stanis{\l}aw and W{\l}odarczyk-Pruszynski, Pawe{\l} and Bengio, Yoshua and Segler, Marwin},
  journal={Journal of Chemical Information and Modeling},
  volume={62},
  number={10},
  pages={2293--2300},
  year={2022},
  publisher={ACS Publications}
}

@article{long2025artificial,
  title={Artificial intelligence in retrosynthesis prediction and its applications in medicinal chemistry},
  author={Long, Lanxin and Li, Rui and Zhang, Jian},
  journal={Journal of Medicinal Chemistry},
  volume={68},
  number={3},
  pages={2333--2355},
  year={2025},
  publisher={ACS Publications}
}

@article{sun2025synllama,
  title={SynLlama: generating synthesizable molecules and their analogs with large language models},
  author={Sun, Kunyang and Bagni, Dorian and Cavanagh, Joseph M and Wang, Yingze and Sawyer, Jacob M and Zhou, Bo and Gritsevskiy, Andrew and Zhang, Oufan and Head-Gordon, Teresa},
  journal={ACS Central Science},
  year={2025},
  publisher={ACS Publications}
}

@article{gao2025synformer,
  title={Generative AI for navigating synthesizable chemical space},
  author={Gao, Wenhao and Luo, Shitong and Coley, Connor W},
  journal={Proceedings of the National Academy of Sciences},
  volume={122},
  number={41},
  pages={e2415665122},
  year={2025},
  publisher={National Academy of Sciences}
}

@inproceedings{luo2024projecting,
  title={Projecting Molecules into Synthesizable Chemical Spaces},
  author={Luo, Shitong and Gao, Wenhao and Wu, Zuofan and Peng, Jian and Coley, Connor W and Ma, Jianzhu},
  booktitle={International Conference on Machine Learning},
  pages={33289--33304},
  year={2024},
  organization={PMLR}
}

@article{lee2025reasyn,
  title={Rethinking molecule synthesizability with chain-of-reaction},
  author={Lee, Seul and Kreis, Karsten and Veccham, Srimukh Prasad and Liu, Meng and Reidenbach, Danny and Paliwal, Saee and Nie, Weili and Vahdat, Arash},
  journal={arXiv preprint arXiv:2509.16084},
  year={2025}
}

@article{gao2021synnet,
  title={Amortized tree generation for bottom-up synthesis planning and synthesizable molecular design},
  author={Gao, Wenhao and Mercado, Roc{\'\i}o and Coley, Connor W},
  journal={arXiv preprint arXiv:2110.06389},
  year={2021}
}

@article{zhu2025meco,
  title={Coder as Editor: Code-driven Interpretable Molecular Optimization},
  author={Zhu, Wenyu and Li, Chengzhu and Tian, Xiaohe and Wang, Yifan and Jia, Yinjun and Wang, Jianhui and Gao, Bowen and Zhang, Ya-Qin and Ma, Wei-Ying and Lan, Yanyan},
  journal={arXiv preprint arXiv:2510.14455},
  year={2025}
}

@article{luo2025prexsyn,
  title={Efficient and Programmable Exploration of Synthesizable Chemical Space},
  author={Luo, Shitong and Coley, Connor W},
  journal={arXiv preprint arXiv:2512.00384},
  year={2025}
}

@article{xie2025transpharmer,
  title={Accelerating discovery of bioactive ligands with pharmacophore-informed generative models},
  author={Xie, Weixin and Zhang, Jianhang and Xie, Qin and Gong, Chaojun and Ren, Yuhao and Xie, Jin and Sun, Qi and Xu, Youjun and Lai, Luhua and Pei, Jianfeng},
  journal={Nature communications},
  volume={16},
  number={1},
  pages={2391},
  year={2025},
  publisher={Nature Publishing Group UK London}
}

@article{gao2025cidd,
  title={Pushing the boundaries of Structure-Based Drug Design through Collaboration with Large Language Models},
  author={Gao, Bowen and Huang, Yanwen and Liu, Yiqiao and Xie, Wenxuan and Ma, Wei-Ying and Zhang, Ya-Qin and Lan, Yanyan},
  journal={arXiv preprint arXiv:2503.01376},
  year={2025}
}

@article{zhao2025chemdfmr,
  title={ChemDFM-R: An chemical reasoner LLM enhanced with atomized chemical knowledge},
  author={Zhao, Zihan and Chen, Bo and Wan, Ziping and Chen, Lu and Lin, Xuanze and Yu, Shiyang and Zhang, Situo and Ma, Da and Zhu, Zichen and Zhang, Danyang and others},
  journal={arXiv preprint arXiv:2507.21990},
  year={2025}
}

@article{ramos2025reviewchemllm,
  title={A review of large language models and autonomous agents in chemistry},
  author={Ramos, Mayk Caldas and Collison, Christopher J and White, Andrew D},
  journal={Chemical science},
  year={2025},
  publisher={Royal Society of Chemistry}
}

@article{zhao2025superchem,
  title={SUPERChem: A Multimodal Reasoning Benchmark in Chemistry},
  author={Zhao, Zehua and Huang, Zhixian and Li, Junren and Lin, Siyu and Zhou, Junting and Cao, Fengqi and Zhou, Kun and Ge, Rui and Long, Tingting and Zhu, Yuexiang and others},
  journal={arXiv preprint arXiv:2512.01274},
  year={2025}
}

@article{ganeeva2025measuringllmsmiles,
  title={Measuring Chemical LLM robustness to molecular representations: a SMILES variation-based framework},
  author={Ganeeva, Veronika and Khrabrov, Kuzma and Kadurin, Artur and Tutubalina, Elena},
  journal={Journal of Cheminformatics},
  volume={17},
  number={1},
  pages={1--15},
  year={2025},
  publisher={Springer}
}

@inproceedings{tao2025makevalidsmiles,
  title={How to Make Large Language Models Generate 100\% Valid Molecules?},
  author={Tao, Wen and Tang, Jing and Chan, Alvin and Hooi, Bryan and Bi, Baolong and Peng, Nanyun and Liu, Yuansheng and Wang, Yiwei},
  booktitle={Proceedings of the 2025 Conference on Empirical Methods in Natural Language Processing},
  pages={26576--26591},
  year={2025}
}

@article{team2023gemini,
  title={Gemini: a family of highly capable multimodal models},
  author={Team, Gemini and Anil, Rohan and Borgeaud, Sebastian and Alayrac, Jean-Baptiste and Yu, Jiahui and Soricut, Radu and Schalkwyk, Johan and Dai, Andrew M and Hauth, Anja and Millican, Katie and others},
  journal={arXiv preprint arXiv:2312.11805},
  year={2023}
}

@article{wei2022cot,
  title={Chain-of-thought prompting elicits reasoning in large language models},
  author={Wei, Jason and Wang, Xuezhi and Schuurmans, Dale and Bosma, Maarten and Xia, Fei and Chi, Ed and Le, Quoc V and Zhou, Denny and others},
  journal={Advances in neural information processing systems},
  volume={35},
  pages={24824--24837},
  year={2022}
}

@article{stumpfe2014recentactivitycliff,
  title={Recent progress in understanding activity cliffs and their utility in medicinal chemistry: miniperspective},
  author={Stumpfe, Dagmar and Hu, Ye and Dimova, Dilyana and Bajorath, Jürgen},
  journal={Journal of medicinal chemistry},
  volume={57},
  number={1},
  pages={18--28},
  year={2014},
  publisher={ACS Publications}
}

@article{stumpfe2012exploringactivitycliff,
  title={Exploring activity cliffs in medicinal chemistry: miniperspective},
  author={Stumpfe, Dagmar and Bajorath, Jurgen},
  journal={Journal of medicinal chemistry},
  volume={55},
  number={7},
  pages={2932--2942},
  year={2012},
  publisher={ACS Publications}
}

@misc{landrum2006rdkit,
  title={RDKit: Open-source cheminformatics},
  author={Landrum, Greg and others},
  year={2006},
  publisher={Zenodo}
}

@article{brown2020fewshotlearning,
  title={Language models are few-shot learners},
  author={Brown, Tom and Mann, Benjamin and Ryder, Nick and Subbiah, Melanie and Kaplan, Jared D and Dhariwal, Prafulla and Neelakantan, Arvind and Shyam, Pranav and Sastry, Girish and Askell, Amanda and others},
  journal={Advances in neural information processing systems},
  volume={33},
  pages={1877--1901},
  year={2020}
}

@article{rogers2010ecfp,
  title={Extended-connectivity fingerprints},
  author={Rogers, David and Hahn, Mathew},
  journal={Journal of chemical information and modeling},
  volume={50},
  number={5},
  pages={742--754},
  year={2010},
  publisher={ACS Publications}
}

@article{mcgregor1999pharm2d,
  title={Pharmacophore fingerprinting. 1. Application to QSAR and focused library design},
  author={McGregor, Malcolm J and Muskal, Steven M},
  journal={Journal of chemical information and computer sciences},
  volume={39},
  number={3},
  pages={569--574},
  year={1999},
  publisher={ACS Publications}
}

@article{baillif2024genbench3d,
  title={Benchmarking structure-based three-dimensional molecular generative models using GenBench3D: ligand conformation quality matters},
  author={Baillif, Benoit and Cole, Jason and McCabe, Patrick and Bender, Andreas},
  journal={arXiv preprint arXiv:2407.04424},
  year={2024}
}

@article{ragoza2022ligan,
  title={Generating 3D molecules conditional on receptor binding sites with deep generative models},
  author={Ragoza, Matthew and Masuda, Tomohide and Koes, David Ryan},
  journal={Chemical science},
  volume={13},
  number={9},
  pages={2701--2713},
  year={2022},
  publisher={Royal Society of Chemistry}
}

@inproceedings{peng2022pocket2mol,
  title={Pocket2mol: Efficient molecular sampling based on 3d protein pockets},
  author={Peng, Xingang and Luo, Shitong and Guan, Jiaqi and Xie, Qi and Peng, Jian and Ma, Jianzhu},
  booktitle={International conference on machine learning},
  pages={17644--17655},
  year={2022},
  organization={PMLR}
}

@article{zhang2023resgen,
  title={ResGen is a pocket-aware 3D molecular generation model based on parallel multiscale modelling},
  author={Zhang, Odin and Zhang, Jintu and Jin, Jieyu and Zhang, Xujun and Hu, RenLing and Shen, Chao and Cao, Hanqun and Du, Hongyan and Kang, Yu and Deng, Yafeng and others},
  journal={Nature Machine Intelligence},
  volume={5},
  number={9},
  pages={1020--1030},
  year={2023},
  publisher={Nature Publishing Group UK London}
}

@article{schneuing2024diffsbdd,
  title={Structure-based drug design with equivariant diffusion models},
  author={Schneuing, Arne and Harris, Charles and Du, Yuanqi and Didi, Kieran and Jamasb, Arian and Igashov, Ilia and Du, Weitao and Gomes, Carla and Blundell, Tom L and Lio, Pietro and others},
  journal={Nature Computational Science},
  volume={4},
  number={12},
  pages={899--909},
  year={2024},
  publisher={Nature Publishing Group US New York}
}

@article{luo20213dsbdd,
  title={A 3D generative model for structure-based drug design},
  author={Luo, Shitong and Guan, Jiaqi and Ma, Jianzhu and Peng, Jian},
  journal={Advances in Neural Information Processing Systems},
  volume={34},
  pages={6229--6239},
  year={2021}
}

@article{trott2010vina,
  title={AutoDock Vina: improving the speed and accuracy of docking with a new scoring function, efficient optimization, and multithreading},
  author={Trott, Oleg and Olson, Arthur J},
  journal={Journal of computational chemistry},
  volume={31},
  number={2},
  pages={455--461},
  year={2010},
  publisher={Wiley Online Library}
}

@article{salentin2015plip,
  title={PLIP: fully automated protein--ligand interaction profiler},
  author={Salentin, Sebastian and Schreiber, Sven and Haupt, V Joachim and Adasme, Melissa F and Schroeder, Michael},
  journal={Nucleic acids research},
  volume={43},
  number={W1},
  pages={W443--W447},
  year={2015},
  publisher={Oxford University Press}
}

@article{schake2025plip2025,
  title={PLIP 2025: introducing protein--protein interactions to the protein--ligand interaction profiler},
  author={Schake, Philipp and Bolz, Sarah Naomi and Linnemann, Katja and Schroeder, Michael},
  journal={Nucleic Acids Research},
  pages={gkaf361},
  year={2025},
  publisher={Oxford University Press}
}

@article{li2024simpretro,
  title={Challenging complexity with simplicity: rethinking the role of single-step models in computer-aided synthesis planning},
  author={Li, Junren and Lin, Kangjie and Pei, Jianfeng and Lai, Luhua},
  journal={Journal of Chemical Information and Modeling},
  volume={64},
  number={14},
  pages={5470--5479},
  year={2024},
  publisher={ACS Publications}
}

@phdthesis{lowe2012uspto,
  title={Extraction of chemical structures and reactions from the literature},
  author={Lowe, Daniel Mark},
  year={2012}
}

@article{lewis2020retrieval,
  title={Retrieval-augmented generation for knowledge-intensive nlp tasks},
  author={Lewis, Patrick and Perez, Ethan and Piktus, Aleksandra and Petroni, Fabio and Karpukhin, Vladimir and Goyal, Naman and K{\"u}ttler, Heinrich and Lewis, Mike and Yih, Wen-tau and Rockt{\"a}schel, Tim and others},
  journal={Advances in neural information processing systems},
  volume={33},
  pages={9459--9474},
  year={2020}
}

@article{li2022ripk,
  title={Generative deep learning enables the discovery of a potent and selective RIPK1 inhibitor},
  author={Li, Yueshan and Zhang, Liting and Wang, Yifei and Zou, Jun and Yang, Ruicheng and Luo, Xinling and Wu, Chengyong and Yang, Wei and Tian, Chenyu and Xu, Haixing and others},
  journal={Nature Communications},
  volume={13},
  number={1},
  pages={6891},
  year={2022},
  publisher={Nature Publishing Group UK London}
}

\end{document}

\maketitle

\tableofcontents
\newpage

\section{Methodological Details}
\label{sec:method}

\subsection{Large Language Model Configuration}
In this study, we utilized the \texttt{gemini-2.5-pro} model (accessed via API) as the reasoning engine. The temperature was set to standard values to balance creativity and adherence to instructions. For the generated outputs, we employed a strict parsing logic to extract the JSON-formatted edit sequences. The maximum number of retries for malformed outputs was set to 3.

\subsection{Definition of the Editing Action Space}
To transform the molecular optimization task into a discrete reasoning problem, we defined a comprehensive action space. The LLM is restricted to outputting a sequence of commands operating on specific atom indices (map numbers). The valid commands and their arguments are defined as follows:

\begin{itemize}
    \item \texttt{["DEL\_ATOM", map\_num]}: Deletes the atom specified by the map number and its incident bonds.
    \item \texttt{["MUTATE\_ATOM", map\_num, new\_symbol]}: Changes the atomic element of the target atom (e.g., C to N).
    \item \texttt{["ADD\_ATOM", new\_id, symbol]}: Introduces a new atom. To avoid index conflict, new IDs start from 500.
    \item \texttt{["DEL\_BOND", map\_num1, map\_num2]}: Removes the bond between two existing atoms.
    \item \texttt{["ADD\_BOND", id1, id2, bond\_type]}: Creates a bond between atoms. Bond types include "SINGLE", "DOUBLE", "TRIPLE", and "AROMATIC".
    \item \texttt{["CHANGE\_BOND", map\_num1, map\_num2, new\_bond\_type]}: Modifies the order of an existing bond.
    \item \texttt{["SET\_CHIRAL", id, tag]}: Explicitly sets stereocenters (CW/CCW).
    \item \texttt{["STOP"]}: A mandatory token indicating the end of the edit sequence.
\end{itemize}

\subsection{Prompt Engineering}

\subsubsection{System Prompt}
The system prompt establishes the persona of the model and defines the output constraints. The full text used in our experiments is provided below:

\begin{lstlisting}[caption={System Prompt used for SynCraft}, label={lst:system_prompt}]
You are an expert medicinal chemist and a computational chemist. Your primary skill is analyzing molecular structures to identify parts that are synthetically challenging and proposing minimal, chemically sound modifications to improve their synthesizability while preserving key pharmacophoric features.

[TASK]
You will be given a source molecule with atom map numbers. Your tasks are:
1. Reason Step-by-Step: First, analyze the molecule to identify synthetically problematic substructures. Then, propose a chemically sound modification.
2. Output Edits: Second, translate your proposed modification into a sequence of machine-readable commands in a JSON list.

[OUTPUT FORMAT]
You MUST provide your response in the following strict format. Do not add any text before '[REASONING]' or after the final '```'.

[REASONING]
Your detailed, step-by-step thinking process goes here. Explain what is wrong with the molecule and why your proposed fix is better.

[EDITS]
```json
[
    ["COMMAND_1", "ARG_1", ...],
    ["STOP"]    
]

[COMMANDS SPECIFICATION]
(The definitions of ADD_ATOM, DEL_ATOM etc. as described in S1.2)
\end{lstlisting}

\subsubsection{Few-Shot Exemplar Construction}
We utilized a Retrieval-Augmented Generation (RAG) approach. For each query molecule, we retrieved the top-k (k=5) most similar examples from our training set based on Morgan Fingerprint similarity. Each exemplar consists of the source SMILES, the ground-truth edit sequence, and a reasoning rationale. An example of a formatted few-shot exemplar is shown below:

\begin{lstlisting}[caption={Example of a Few-Shot Exemplar}, label={lst:few_shot}]
source_smiles: "[CH3:1][NH:2][CH2:3][C@H:4]1[CH2:5][CH2:6]N:7[CH2:11]1"

[REASONING]
The N-acetyl group on the pyrrolidine nitrogen presents a potential synthetic liability due to the acidity of its alpha-protons, which can lead to unwanted enolization under basic conditions. The transformation replaces the methyl of the acetyl group with a tert-butyl group, effectively converting the N-acetyl moiety into a more sterically hindered N-pivaloyl group. This is an intelligent modification to enhance synthetic utility and metabolic stability.

[EDITS]
```json
[
  ["ADD_ATOM", 500, "C"],
  ["ADD_ATOM", 501, "C"],
  ["ADD_BOND", 9, 500, "SINGLE"],
  ["ADD_BOND", 9, 501, "SINGLE"],
  ["STOP"]
]```

\end{lstlisting}
\subsection{Interaction-Aware Constraints Pipeline}
For the structure-based optimization tasks, we implemented an automated pipeline to extract biological constraints.
\begin{enumerate}
\item \textbf{Docking:} The input SMILES is converted to a 3D conformer and docked into the reference protein structure using AutoDock Vina.
\item \textbf{Interaction Profiling:} The best-scoring pose is analyzed using the Protein-Ligand Interaction Profiler (PLIP). We detect Hydrogen Bonds, Hydrophobic Interactions, Salt Bridges, Pi-Stacking, Pi-Cation, Halogen Bonds, and Metal Complexes.
\item \textbf{Graph Alignment:} A substructure matching algorithm maps the PDB atom serial numbers from the docked pose back to the canonical atom map numbers of the 2D molecular graph. This ensures the constraints correspond correctly to the atoms in the edit action space.
\item \textbf{Constraint Injection:} The detected interactions are converted into natural language descriptions and appended to the user prompt.
\end{enumerate}
The format of the injected constraint prompt is:
\begin{lstlisting}
CRITICAL BIOLOGICAL CONSTRAINTS:
The following atoms from the source molecule are critical for binding to the target protein and you should be careful when modifying them...
Atom with map number [12] (a N atom connected to [C:11, C:13]) forms a critical Hydrogen Bond with ASP156.
Atom with map number [23] (a C atom connected to [C:22]) is involved in a key Hydrophobic interaction with LEU145.
\end{lstlisting}
\newpage
\section{Ablation Studies}
\label{sec:ablation}

To validate the architectural choices of SynCraft, we conducted a comprehensive ablation study focusing on two critical components: the output modality (Edit Sequence Prediction vs. Direct SMILES Generation) and the inference strategy (Few-shot Retrieval-Augmented Generation vs. Zero-shot).

\subsection{Experimental Setup}
We evaluated four different configurations on the test sets derived from Pocket2Mol (P2M) and ResGen:
\begin{enumerate}
    \item \textbf{5-shot with edits (SynCraft Standard):} The proposed method using 5 retrieved exemplars to predict edit sequences.
    \item \textbf{1-shot with edits:} Using only the single nearest neighbor as a demonstration to predict edit sequences.
    \item \textbf{5-shot direct-SMILES:} Using 5 retrieved exemplars but instructing the LLM to directly generate the full SMILES string of the optimized molecule.
    \item \textbf{Zero-shot direct-SMILES:} Asking the LLM to optimize the molecule directly without any in-context examples.
\end{enumerate}

The primary metric is the success rate of identifying a synthesizable analog that maintains a Tanimoto similarity above specific thresholds ($0.5$ to $0.9$) relative to the original input.

\subsection{Results}

The quantitative results on the Pocket2Mol and ResGen datasets are presented in Table \ref{tab:ablation_p2m} and Table \ref{tab:ablation_resgen}, respectively.

\begin{table}[H]
\centering
\caption{\textbf{Ablation Study on Pocket2Mol (P2M) Dataset.} Success rates of finding synthesizable analogs at varying similarity thresholds. The ``Total Solved" column indicates the percentage of inputs for which a valid, synthesizable molecule was generated, regardless of similarity.}
\label{tab:ablation_p2m}
\resizebox{\textwidth}{!}{%
\begin{tabular}{lcccccc}
\toprule
\textbf{Configuration} & \textbf{Total Solved} & \textbf{Sim $>$ 0.5} & \textbf{Sim $>$ 0.6} & \textbf{Sim $>$ 0.7} & \textbf{Sim $>$ 0.8} & \textbf{Sim $>$ 0.9} \\
\midrule
\textbf{5-shot with edits (SynCraft)} & 61.34\% & \textbf{38.54\%} & \textbf{25.71\%} & \textbf{10.90\%} & \textbf{3.09\%} & 0.47\% \\
1-shot with edits & 50.73\% & 33.24\% & 21.57\% & 9.50\% & 2.92\% & \textbf{0.52\%} \\
5-shot direct-SMILES & 59.77\% & 31.66\% & 19.71\% & 7.64\% & 2.10\% & 0.29\% \\
Zero-shot direct-SMILES & \textbf{67.35\%} & 23.62\% & 11.55\% & 4.31\% & 1.17\% & 0.29\% \\
\bottomrule
\end{tabular}%
}
\end{table}

\begin{table}[H]
\centering
\caption{\textbf{Ablation Study on ResGen Dataset.} Success rates of finding synthesizable analogs at varying similarity thresholds.}
\label{tab:ablation_resgen}
\resizebox{\textwidth}{!}{%
\begin{tabular}{lcccccc}
\toprule
\textbf{Configuration} & \textbf{Total Solved} & \textbf{Sim $>$ 0.5} & \textbf{Sim $>$ 0.6} & \textbf{Sim $>$ 0.7} & \textbf{Sim $>$ 0.8} & \textbf{Sim $>$ 0.9} \\
\midrule
\textbf{5-shot with edits (SynCraft)} & 69.27\% & \textbf{44.68\%} & \textbf{29.07\%} & \textbf{13.07\%} & \textbf{3.51\%} & 0.49\% \\
1-shot with edits & 59.32\% & 38.93\% & 25.85\% & 10.63\% & 2.34\% & 0.20\% \\
5-shot direct-SMILES & 64.20\% & 34.83\% & 21.76\% & 8.39\% & 2.05\% & \textbf{0.78\%} \\
Zero-shot direct-SMILES & \textbf{72.39\%} & 26.44\% & 15.22\% & 5.66\% & 1.56\% & 0.39\% \\
\bottomrule
\end{tabular}%
}
\end{table}

\subsection{Discussion}

\subsubsection{Superiority of Edit Prediction over Direct Generation}
A key finding from the tables is the distinct behavior between editing and direct generation. While the \textit{Zero-shot direct-SMILES} baseline achieves the highest ``Total Solved" rate (67.35\% for P2M and 72.39\% for ResGen), its performance drops precipitously when a similarity constraint is applied. For instance, on ResGen, the success rate for Zero-shot drops from 72.39\% (any molecule) to 26.44\% (Sim $>0.5$), whereas SynCraft maintains a robust 44.68\%.

This phenomenon indicates that without the structural constraints imposed by the edit-based paradigm, LLMs tend to ``hallucinate" simple, generic synthesizable molecules that bear little resemblance to the complex input scaffold. By predicting discrete edit operations (e.g., \texttt{DEL\_ATOM}, \texttt{ADD\_BOND}), SynCraft forces the optimization to occur \textit{on the graph of the original molecule}. This inherently preserves the majority of the structural features, resulting in significantly higher success rates at medium-to-high similarity thresholds ($>0.6$), which is the critical regime for lead optimization.

Comparing the ``5-shot with edits" against ``5-shot direct-SMILES" further confirms this. Even when provided with the exact same few-shot examples, the edit-based output modality consistently outperforms the SMILES generation modality (e.g., 29.07\% vs 21.76\% at Sim $>0.6$ on ResGen). This validates our hypothesis that LLMs are better reasoned as ``chemical editors" rather than ``chemical translators" for this task.

\subsubsection{Importance of Few-Shot Retrieval (RAG)}
The comparison between 5-shot, 1-shot, and Zero-shot configurations highlights the necessity of in-context learning.
\begin{itemize}
    \item \textbf{Contextual Grounding:} The ``5-shot" configuration consistently outperforms the ``1-shot" configuration, suggesting that providing diverse examples of valid chemical transformations helps the model generalize better to unseen scaffolds.
    \item \textbf{Guidance against Mode Collapse:} The poor performance of the Zero-shot baseline at high similarity thresholds suggests that without retrieving relevant ``Synthesis Cliff" pairs, the model lacks the specific knowledge required to perform surgical repairs. Instead, it defaults to generating training-set-like synthesizable molecules, leading to the ``structural drift" observed in the data.
\end{itemize}

In summary, the ablation study provides strong empirical evidence that the combination of \textbf{generative editing} and \textbf{retrieval-augmented inference} is essential for solving the synthesizability optimization problem effectively.

\newpage

\section{Prospective Optimization Results for All 42 High-Scoring RIPK1 Candidates}

To provide a comprehensive and unbiased assessment of SynCraft’s utility in real-world lead optimization scenarios, we report the complete optimization results for the 42 high-scoring RIPK1 candidates described in the main text.
These candidates were originally identified from a generative library targeting the RIPK1 kinase but were discarded due to structural complexity or lack of viable synthetic pathways. SynCraft was tasked with optimizing these molecules using the interaction-aware prompting strategy (PDB ID: 7YDX) to restore synthesizability while preserving predicted binding affinity.
\begin{figure}
    \centering
    \includegraphics[width=\linewidth]{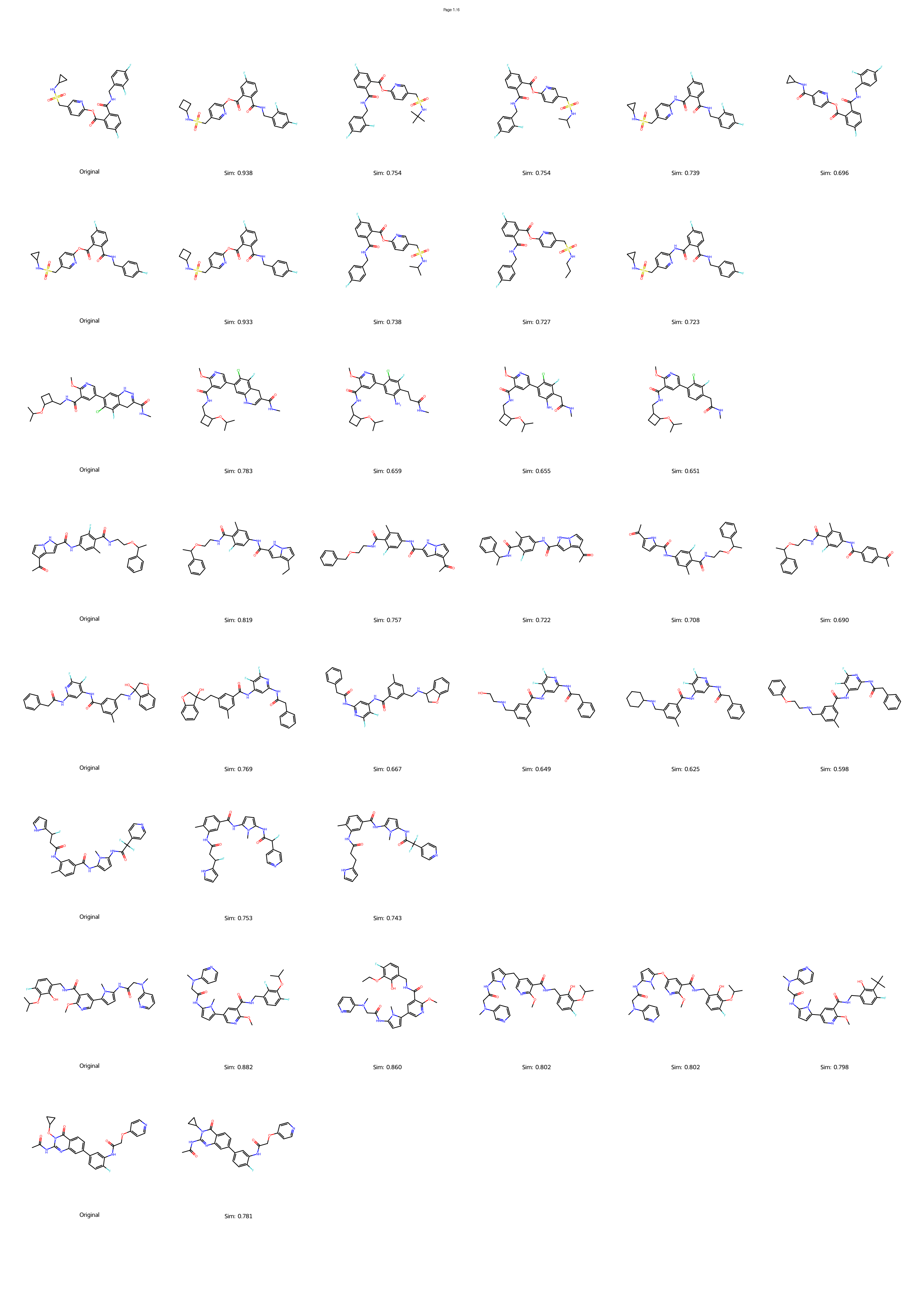}
\end{figure}

\begin{figure}
    \centering
    \includegraphics[width=\linewidth]{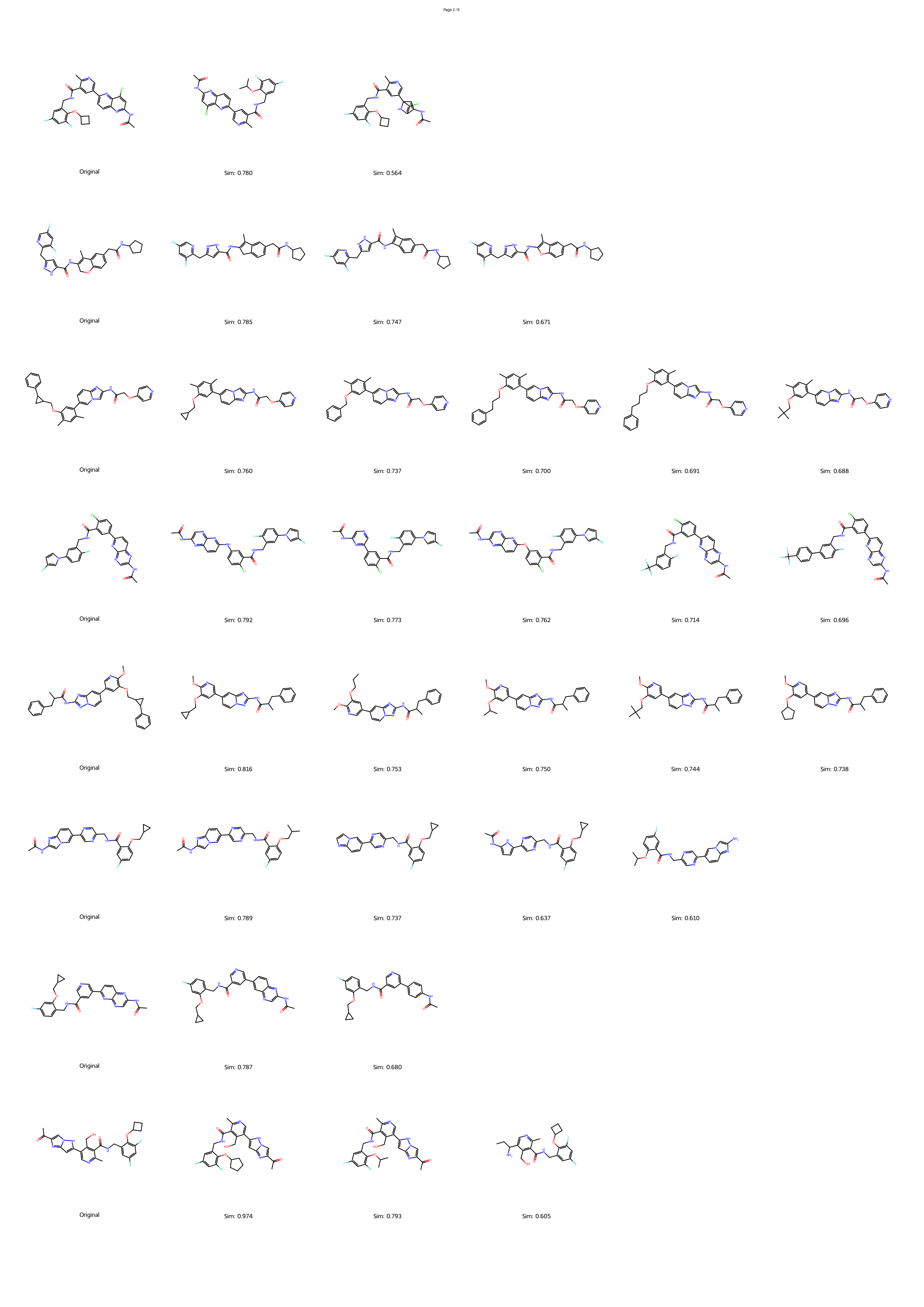}
\end{figure}

\begin{figure}
    \centering
    \includegraphics[width=\linewidth]{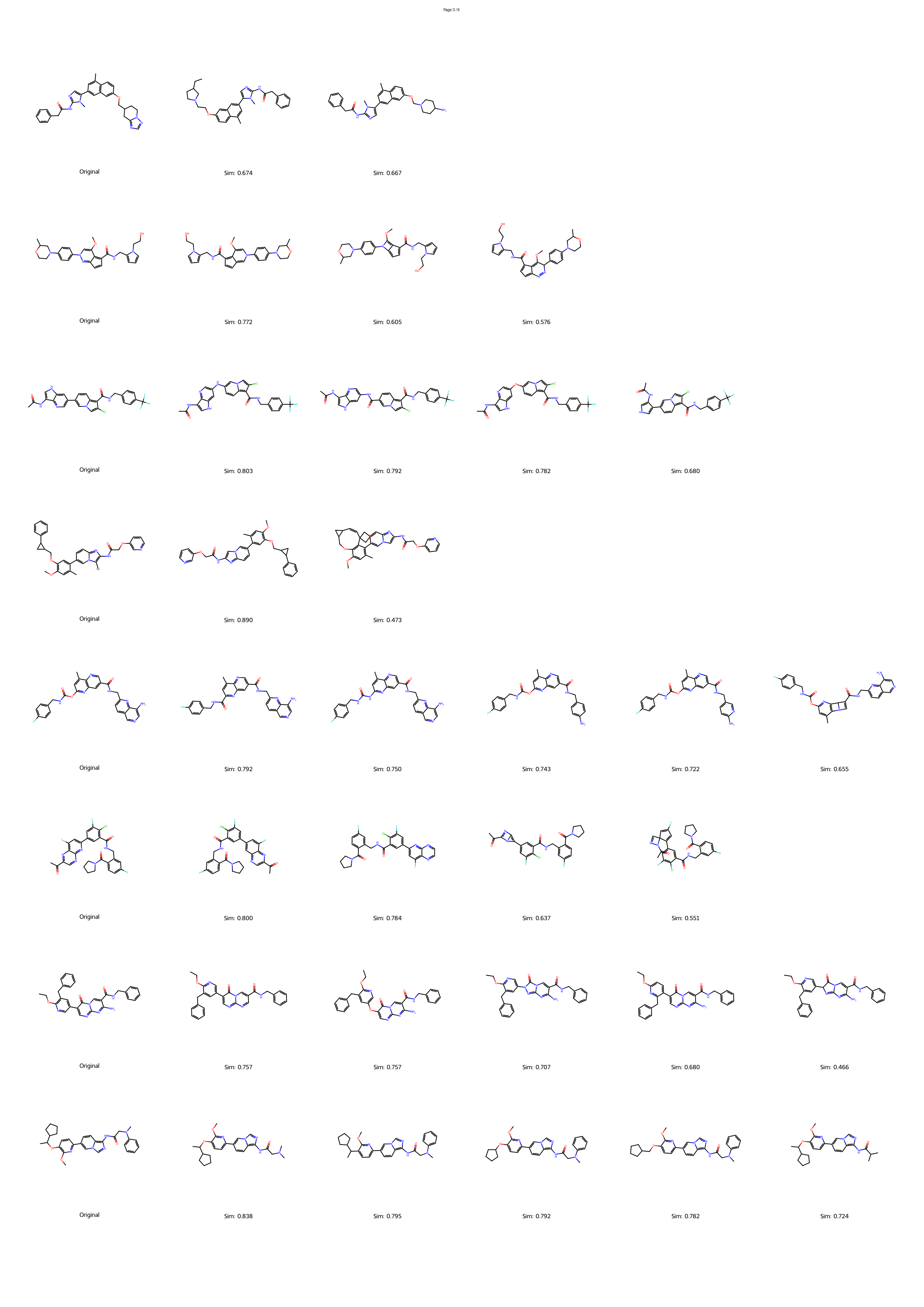}
\end{figure}

\begin{figure}
    \centering
    \includegraphics[width=\linewidth]{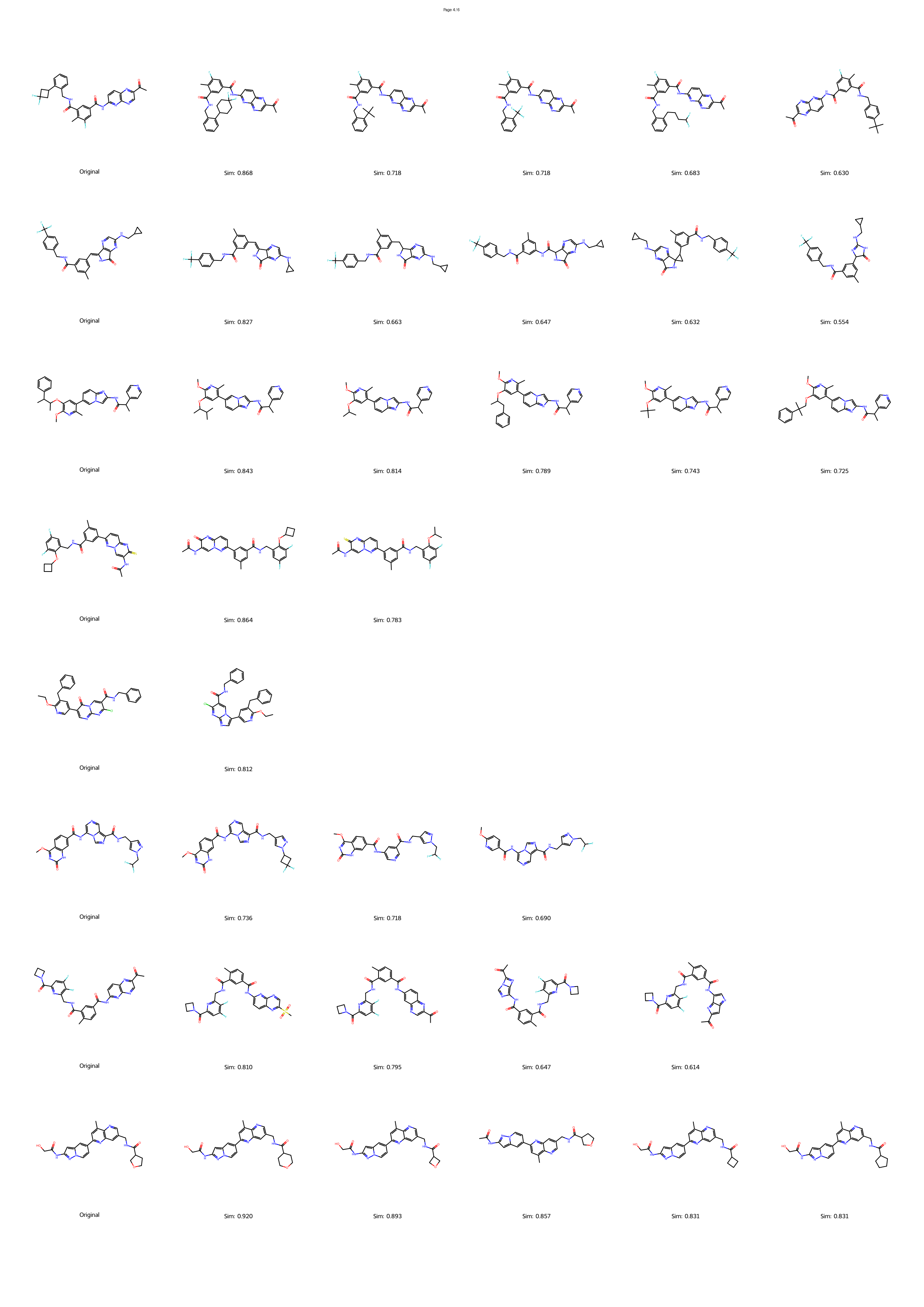}
\end{figure}

\begin{figure}
    \centering
    \includegraphics[width=\linewidth]{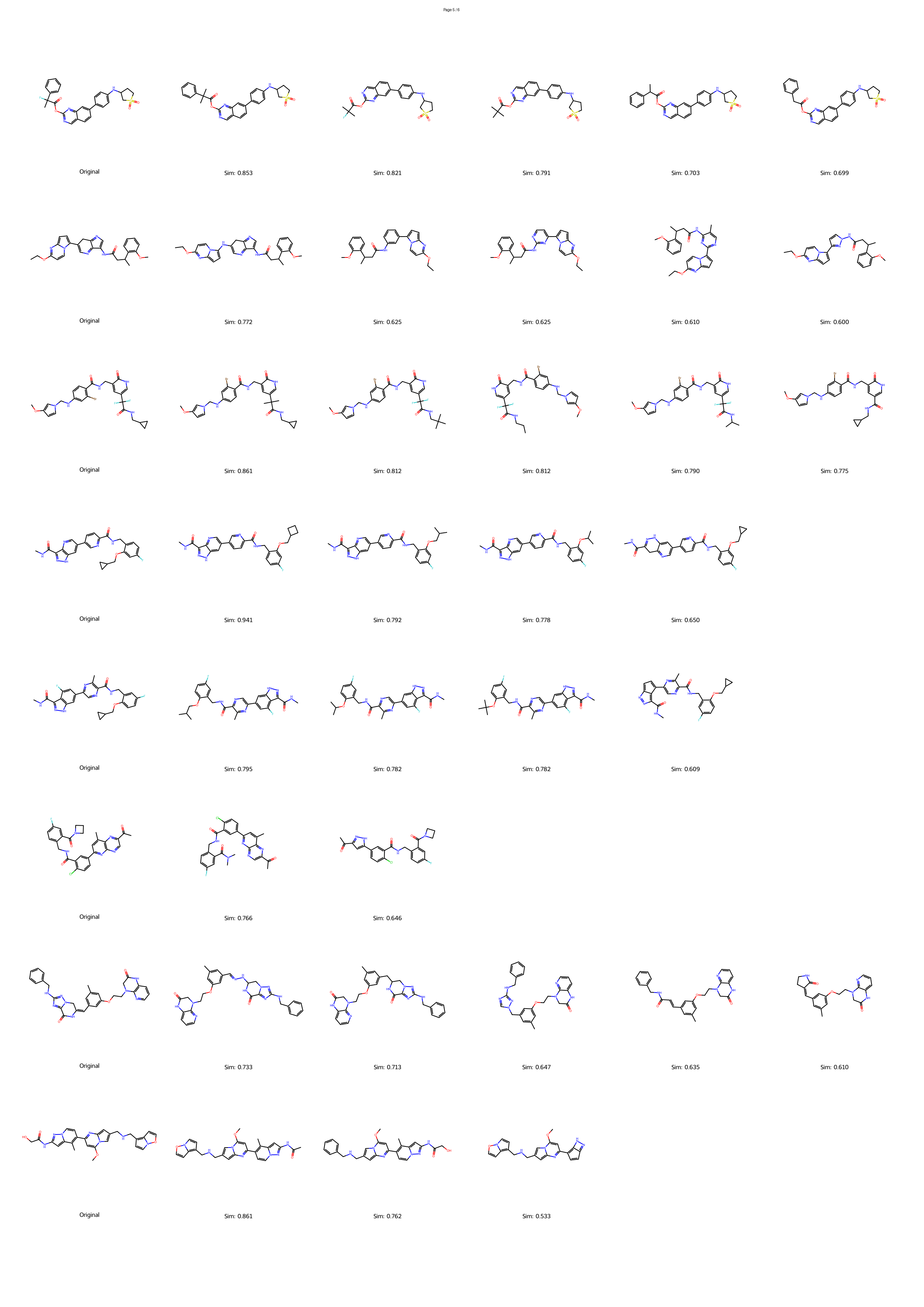}
\end{figure}

\begin{figure}
    \centering
    \includegraphics[width=\linewidth]{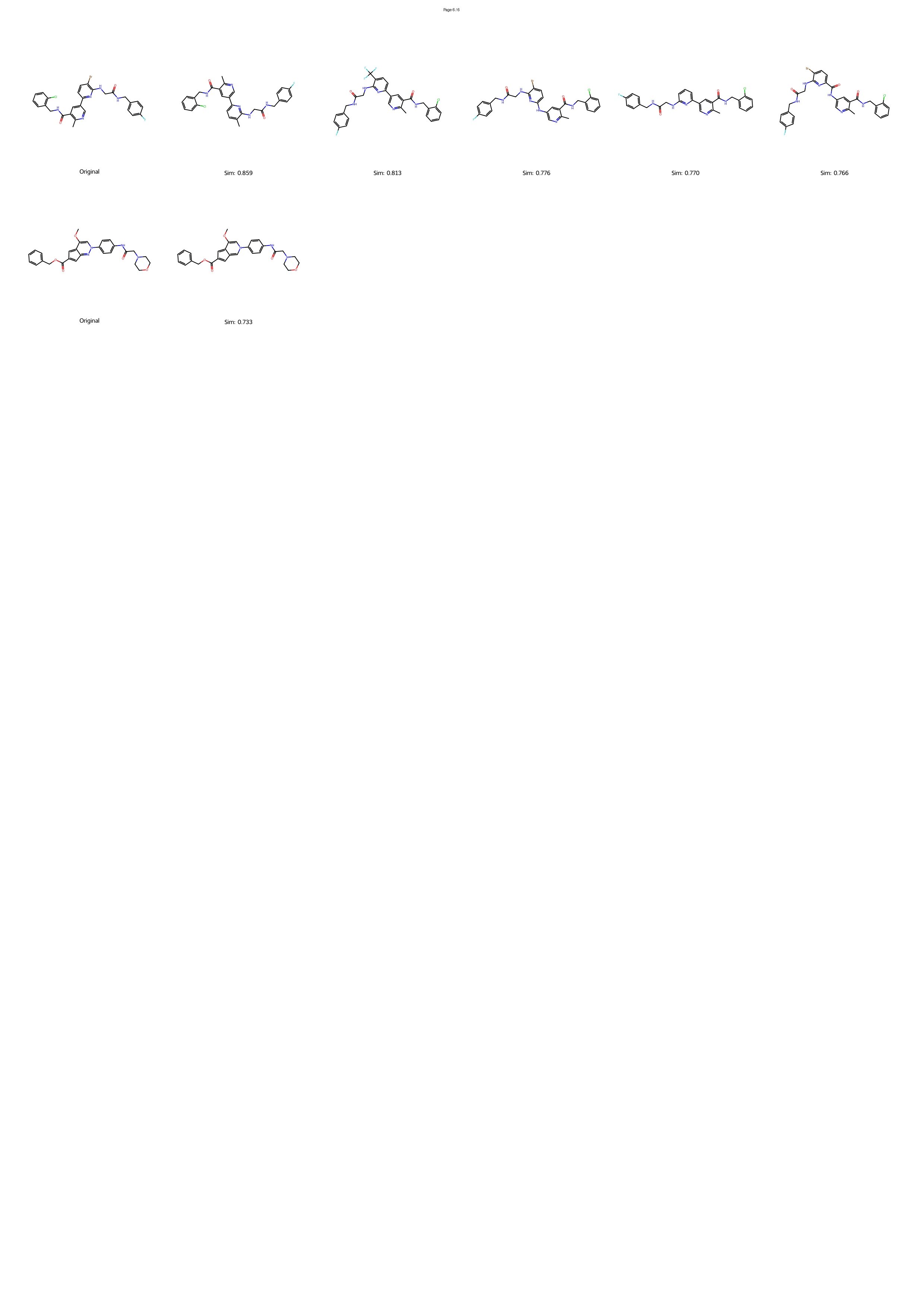}
\end{figure}

\bibliography{achemso-demo}